\documentclass{article}

\usepackage{hhline}
\usepackage{subcaption}
\usepackage{colortbl}
\usepackage{wrapfig}
\usepackage{ccicons}  
\usepackage{tcolorbox}
 
\definecolor{Gray}{gray}{0.85}
\definecolor{LightCyan}{rgb}{0.88,1,1}
\newcolumntype{a}{>{\columncolor{Gray}}p{1.8cm}}
\newcolumntype{b}{>{\columncolor{Gray}}p{3cm}}
\newcolumntype{c}{>{\columncolor{Gray}}p{4cm}}

\usepackage{natbib}
\usepackage{hyperref}
 \usepackage{authblk}
 \usepackage{enumitem}
\begin{document}

\title{
Automated detection of dark patterns in cookie banners: how to do it poorly and why it is hard to do it any other way}

\author[1]{Than Htut Soe}
\author[2]{Cristiana Teixeira Santos}
\author[1]{Marija Slavkovik}
 
\affil[1]{University of Bergen, Norway,\{than.soe, marija.slavkovik\}@uib.no}
\affil[2]{Utrecht University, The Netherlands, c.teixeirasantos@uu.nl \\}
\date{}                 
\maketitle



\begin{abstract}
Cookie banners, the pop ups that appear to collect your consent for data collection,  
are a tempting ground for dark patterns.  
Dark patterns are design elements that are used  
to influence the user's choice  towards an option that is not in their interest. The use of dark patterns renders consent elicitation meaningless and voids the attempts to improve  
a fair collection and use of data.  
Can  machine learning be used to automatically detect the presence of dark patterns in cookie banners? 
In this work, a dataset of cookie banners  
of 300 news websites was used to train a prediction model that does  
exactly that.  
The machine learning pipeline we used includes feature engineering, parameter search, training a Gradient Boosted Tree classifier and evaluation. 
The accuracy of the trained model is promising, but allows a lot of room for improvement.
We provide an  
in-depth analysis of the interdisciplinary challenges that automated dark pattern detection poses to artificial intelligence. 
 The dataset and all the code created using machine learning is available at the url to repository removed for review.  
\end{abstract}

\section{Introduction}

The digitization of our daily lives has ushered opportunities for the collection of personal data on those activities, which up to recently, were private.  Moreover, 
automated data collection has become almost impossible to escape \citep{Auxier2019}.  This  data is used in ways that can impact our reality \citep{aies-2021}.
Regulations are increasingly put in place to protect our interests. Examples include the 
European Union's General Data Protection Regulation (GDPR)~\citep{GDPR} 
and the California Consumer Privacy Act (CCPA) \citep{CCPA}. 
One of the goals of these regulations is 
to enforce the practice to inform and obtain consent from users about the use of data processing and data trackers \citep{utz_informed_2019}. 
In Europe, the most tangible effect of these regulations has been the appearance, and ubiquity, of consent banners on webpages.   Cookie banners appear because, according to the GDPR and the ePrivacy Directive (ePD)~\citep{ePD-09}, websites, regardless of where they are based, must inform users located in the EU about personal data collection and obtain their consent for certain purposes\footnote{An example of such cookie banner can be seen in  Figure~\ref{fig:cookieconsent} of the Appendix \ref{APPENDIX}.}.

%
 
As awareness of data collection and its effects increases, so does the complexity and variety of cookie banners throughout websites and different modalities.
Website publishers and their designers, driven by business, marketing  needs~\citep{Gray-chivukula-mediationUX, chivukula-design}, design  user interfaces (UI)  
deliberately make use of the breadth  space  they are afforded with regarding UI design.
\emph{UI design} has not been explicitly nor extensively specified in mandatory regulations nor guidelines regarding consent~\citep{Sant-etal-20-TechReg, Karegar-et-al}.
This gap allows, as several studies have already shown, ~\citep{utz_informed_2019,MatteBS2019,NouwensLVKK20,SoeNGS20,Grassl2021,bauer_are_2021, gray-legalreq}, 
to make use of (weaponize~\citep{WALDMAN2020105})  the interface design  to steer and manipulate users towards privacy choices they would not normally have an incentive to take.
Such interface design is called ``{\em dark pattern}''used to circumvent the user's genuine choice and  the intent of privacy regulations~\citep{ForbrukerradeNO,CNIL-Shap-19}, even when those are explicitly required to protect users, as is the case with cookie banners 
\citep{DucatoE2019}.
The volume  and pervasiveness of questionable cookie banners embedding dark patterns 
 surpasses any human capacity to detect, report and penalize violations for non-compliant practices, either in at desktop web, mobile web and mobile app~\citep{Gunawan-comparativestudy2021}. 
 Several  studies measured the presence and
behavior of cookie banners on major websites and found that well over 50\% contained dark
patterns~\citep{nouwens_dark_2020, Sanchez-Rola-cookies2019, utz_informed_2019, SoeNGS20, Soheil2021, matte_cookie_2020}, e.g. when banners present different colors, sizes or shapes of options (violating the  unambiguous consent requirement).
%

Any regulation only constitutes a strong deterrent against any illegal practice, if such regulation can be efficiently \emph{enforced} at court or regulatory level. 
Yet, most consent banners embedding dark patterns go undetected and unsanctioned, as data protection regulator’s IT resources are largely insufficient to address all the suspected legal breaches~\citep{AccessNow-2020,brave-2020}. This reality casts doubts on their real possibilities to investigate these matters efficiently, promptly and on a large scale. Thus, there is a need for \textit{technical} solutions (algorithms and prototypes) to collect  and analyze reliable data on dark patterns, expedite oversight tasks, warn and protect users, expose manipulative practices, and provide proof of unlawful influence to support legal proceedings.
Clearly, 
\emph{automatic detection} of the presence of dark patterns at scale is pertinent.
However, dark pattern detection is a complex cognitive task which makes 
the use of artificial intelligence (AI) particularly challenging. 

In this paper, we are concerned with this problem of using AI, specifically supervised machine learning (ML),  for automating the detection of dark patterns in 
cookie banners. 
We present an initial approach. 
We used a rich manually labeled dataset made available by \citet{SoeNGS20} and supervised ML to train several prediction models that identify whether or not a cookie banner has a dark pattern. In particular, a cookie banner is represented by a data point. 
Herewith, we consider a data point to be a set of features of different data type values describing the position of the interface on the screen, amount of text, options given to the user, etc. 
Each data point is labeled with the following information: whether it has, possibly has, or is confirmed to have one of the 5 dark patterns categories defined by \citet{gray_dark_2018}: nagging, obstruction, sneaking, interface interference, and forced action.  Thus we have 15 possible different labels. 
A prediction model we trained assigns one of these 15 possible labels to a new data point (the interface represented as a set of features). 

We report the following findings in our experiment.
What we present is 
not a practical approach to detecting dark patterns: our approach relies on being able to encode an interface as a set of feature values before it is fed into the prediction model for dark pattern detection.  
To obtain values for some features, we relied on supervised and unsupervised machine learning. 
Some of these used features are such that their values are easy to harvest automatically, but others require human intervention instead. 
If humans are already looking at the interface, it is most efficient that they are directly tasked with detecting if a dark pattern is present. One can, of course,  argue that laymen users cannot necessarily detect a dark pattern and we could still spare the resources for training human dark pattern detectors. 
%
Lastly, the accuracy of the predictions we obtained is not particularly high. What we present can therefore be understood as a negative result, but it should still be seen as an advancement towards making automated dark pattern detection a reality - we have learned what does work and particularly 
why does not work.    

We discuss why attempts at automated dark patterns detection such as ours are still insufficient. While our attempt exposes many limitations to the automated detection of dark patterns, it also offers valuable lessons on how we can make progress. Towards this goal, we have made our code, prediction models and data  available for anyone to build upon our efforts.
We argue that the automatic detection of dark patterns requires not only machine learning, but also refinement of the concept of dark patterns and consequently  improvements in regulatory initiatives.




\section{Why is  dark patterns detection difficult for AI?}\label{sec:whydifficult}
The term dark patterns \citep{gray_dark_2018} has been coined\footnote{The neologism, dark pattern, was coined by user experience designer Harry Brignull in 2010.} to identify ``instances where designers use their knowledge of human behavior (e.g., psychology) and the desires of end users to implement deceptive functionality that is not in the user's best interest'' \citep{gray_dark_2018}. Throughout this work, we use the concept ``dark pattern''
to refer to types of  UI dark designs that have been documented and applied to cookie banners on websites~\citep{gray_dark_2018, nouwens_dark_2020}.

There is a growing concern that dark patterns can and will be used to i) impede 
ethical artificial intelligence systems by 
hampering the explainability and transparency of such systems \citep{Chromik2019}, and  to ii)  manipulate users into sharing more data with a service than the service needs to operate \citep{Bosch2016}. 
In this section, we decompose some of the challenges of automatic detection of dark patterns by AI methods. 

\subsection{Representation challenges}

Artificial intelligence (AI), specifically machine learning, has made considerable breakthroughs in image recognition and natural language processing (NLP)~\citep{BengioLH21}.  Despite this progress, AI is deployed successfully when the different cognitive tasks are ``emulated'' in isolation: image recognition as one task, language processing and sentiment analysis as separate tasks. However, humans perceive a cookie banner, and any other interface, as an single visual-language experience. 
A dark pattern deceives by forcing the where the user places their attention. 

For any algorithm to be able to process any kind of information, that information needs to be represented in a form that can be handled by the algorithm. For  cookie banner interfaces, we have 3 \emph{representation choices}: as image, as text, or as a described  phenomenon (a.k.a. factorised representation). We discuss the limitations and advantages of each of these three options.

\noindent
\textbf{Images.} 
The input to the algorithm will be the pixels of, effectively,  the screenshot of a screen image  with the interface active on it.  

\noindent
\textit{Advantage.} The advantage of using images is that they are easy to collect automatically -- a user can easily submit a screenshot and signal whether the image contains a dark pattern or not. 
A neural network can be trained using such examples of images with and without dark patterns. However, what neural network  detects in images is the existence of correlation between pixels. 

\noindent
\textit{Limitation.}
The image representation disregards   information from the interface. 
Dark patterns, such as nagging, exist as an event (one image following another) and would be virtually impossible to capture in an image alone. Information within the text of the interface will also be disregarded. Two images with different text can easily end up being considered similar by a neural network even if one contains a dark pattern and the other does not. 
Neural networks can be expected to be successful in identifying how much of the screen does the cookie banner take up, or whether the accept and reject options in a cookie banner are implemented as the same widget. 
Nevertheless, it is worthwhile empirically verifying the limits of image recognition on the task of identifying visual elements of an interface. 


\noindent
\textbf{Text.} 
The input of the algorithm will be natural language text, or a sequence of texts and treated as a sentiment analysis problem \citep{ChaturvediPC18}. 

\noindent
\textit{Advantage.} 
NLP techniques (e.g., sentiment analysis) can help to recognize dark patterns that play on linguistic features (e.g., confirmshaming, tricky questions, toying with emotion, arguments of authority, fear mongering dark patterns)~\citep{kampanos2021accept}.
What would considerably further the abilities of an AI algorithm to detect dark patterns is the legal requirement that privacy options and its text (e.g. accept, reject, configure) should be balanced (or equitable) (Article 7(4) of the GDPR, further interpreted by the data protection community~\citep{Adv-gen-Szpunar-2019,CNIL-draft-rec-cookies-2020, Sant-etal-20-TechReg}). As so, it possible to automatically detect whether two antonym choices are present in the panel.

\noindent
\textit{Limitation.}
Text, compared to image, is harder to automatically scrap from online applications, particularly if the interface interaction invokes other interfaces. 
Cookie banners are observed to employ legal and technical jargon \citep{STRYCHARZ2021106750, utz_informed_2019}, vague and ambiguous language~\citep{Santos2017DetectingAE}, and positive or negative framing~\citep{Hausner2021} which hampers automatic detection of textual expressions.
An open question is whether the text of the interface is sufficient to identify a dark pattern. 
A study that could verify or refute a positive answer to this question would be to compare the insights of one group of study participants tasked to detect dark patterns in an image of an interface, and another group tasked with detecting the dark pattern when only given the text of the interface. Such a study is outside of the scope of this work. 
Necessarily to denote, however, is the fact that only considering text eliminates the possibility to take into account other impacting visual 
cues, such as using different colors or salient hues or values representing different privacy options afforded to the user.

\vspace{1mm} 

\noindent
\textbf{Features.} An option  is to identify a set of discerning features of the interface and record the values of those features. This option is the one  we 
used in this paper. A supervised learning algorithm can  be used to build a prediction model essentially finding a pattern in the feature values that classifies an interface as either containing or not dark patterns and identifying which one. 

\noindent
\textit{Advantage.} Building on the work of~\citet{SoeNGS20}, a large selection of  discerning features is used to describe the properties of online cookie banners from news outlets.   
The advantage of this approach over the other two is that interactive phenomena, such as invoking several interfaces, can be described and  both visual and textual 
cues can be captured in the representation. 
In Section~\ref{sec:methodology} we describe how the features from the  \citet{SoeNGS20} dataset were processed before being used to train a prediction model. 

\noindent
\textit{Limitation.} 
The clear disadvantage is that some feature values will be difficult to be automatically ``harvested'' and virtually all feature values will require different AI post-processing techniques. Namely, after one algorithm (or a human) has identified and scraped the text, another is needed to analyse it. For example, text would require sentiment analysis, and the identification of widgets used would require a image processing. 
%
Instead, an ideal approach, and one open problem for future work consists in identifying the features whose values can be automatically  scraped and post-processed and which are most relevant to distinguish a dark pattern. The more precise the definition of  dark patterns, the easier it is to identify the relevant features.  
A separate limitation of the features approach is the challenge of humans being able to correctly label the data points with the right dark pattern. We discuss this issue next. 

\subsection{Detection challenges}

\noindent
\textbf{Challenges to detect UI of cookie banners}
The GDPR does not address UI-based elements (the same colors assigned for options, position, design, size, format, location, text, etc.).
There is very limited  case-law in the EU concerning the use of UI, which refers mainly to the prohibition to use pre-checked boxes~\citep{CJEU-Planet49-19}.
Non-mandatory guidelines from Data Protection Authorities provide further interpretation on UI elements, wherein feature parity is is given priority to.
The UK DPA \cite[p. 32]{ICO-Guid-19} observed that \emph{“a consent mechanism that emphasises ‘agree’ or ‘allow’ over ‘reject’ or ‘block’ represents a non-compliant approach, as the online service is influencing users towards the ‘accept’ option.”}
The French DPA~\cite[p. 28]{CNIL-Shap-19} frames as  \emph{“Attention Diversion”} the design choices that draw attention to a point of the site or screen to distract or divert the user from other points that could be useful. This guidance states that visual and color saliency is effective and commonly used, indicating that using a green hue on a “continue” button while leaving the “find out more” or “configure” button smaller or in a lighter shade of grey, users may perceive green as the preferable choice. 
Conversely, in the US, the Congress~\citep{Stigler-report-2019} is considering issuing legislation
restricting dark patterns, and the CCPA~\citep{CCPA} 
defines and prohibits dark patterns associated with privacy consent which have a ``substantial effect of subverting or impairing a consumer’s choice to opt-out''.

\vspace{1mm} 

\noindent
\textbf{Challenges on  a consensual definition of dark patterns.}
What exactly makes a pattern \emph{dark}\footnote{A better term would be {\em obscure}.} is still a matter of fervent discussion across different communities~\citep{Mathur-2021}. 
There is a body of theoretical conceptualization on the definition of dark patterns -- researchers unfold diverging definitions and classifications thereof~\citep{mathur_dark_2019, brignull_dark_2015, Bosch2016, CNIL-Shap-19, Chromik2019, gray_dark_2018, Zagal2013DarkPI,Fritsch2017} -- and their related features -- in their own domains, thus even rendering it difficult to communicate about the same phenomenon. 
Pertinent features that determine what makes a certain design “dark”:
nudging~\citep{Acquisti-nudging}, intention, manipulation, influence, persuasion~\citep{Cialdini-persuasion}, deception~\citep{Bongard2021manipulated}, harm (privacy, financial, time, etc., and evidence thereof) are interpreted with great latitude by different disciplines  (law, computer science, cybersecurity, philosophy, ethics) and from different perspectives (e.g., user, designer, developer, lawyer, website publisher, marketer~\citep{gray-legalreq}) 
 and need to be defined with consensus to build dark patterns detection applications.

\vspace{1mm} 

\noindent
\textbf{Challenges to detect intention and deception.}
Besides UI-based elements (e.g. feature inequality), there is also the question of whether AI can capture \textit{intention} and  \textit{deception} in cookie banners. 
If we define dark pattern's existence as an attempt for deception, we need to detect whether and when  intention to deceive exists. 
Existence of intention is very difficult to prove and it is hard both for humans and for machines. 

\vspace{1mm} 

\noindent
\textbf{Challenges to detect contextual, social and 
cognitive aspects.}
Another detection problem refers for context awareness (e.g. temporal, social, cognitive aspects) which are hard to capture. 
 The ability to process all the contextual information at once,  holistically remains a hard problem in AI.
 As of today, AI methods process different types of information differently: for example, text is different than image processing (as mentioned in section 3.1).  
%
%
%
%
Within the UI there is a richness of information that is transmitted to the user that needs to be accounted to: the placement of the interface in the screen, the ratio of the interface size vs the screen size, the contrast between different colors used, the finer linguistic nuances of the text used, etc. Herein, the inputs of the HCI community is of essence. 
%


\subsection{Summary}
Ultimately, dark pattern detection is difficult for AI because it is difficult also for people. It is difficult to capture all the specific instances in which a particular design choice constitutes a dark pattern. The analysis of ~\cite{SoeNGS20}  reveals that reviewers struggled to agree upon on which dark pattern is present in a cookie banner -- more than option applied, given a low  inter-reviewer reliability. A better, more context specific, definition can contribute to at least eliminate this human labeling uncertainty problem. 
A common vocabulary for the identification, description and categorization of dark patterns (and in concrete contexts) is needed for its comprehensive detection.


\section{Methodology}\label{sec:methodology}
All of the machine learning tasks, except word embedding, was done using the Scikit-learn library\footnote{\url{https://scikit-learn.org}} on a Jupyter notebook. Since the word embedding with Universal Sentence encoder and clustering is more computationally intensive, the task was performed with Tensor Flow\footnote{\url{https://www.tensorflow.org/}} library on the Google Collaboratory\footnote{\url{https://colab.research.google.com}}, a cloud computational platform.

\subsection{Dataset} 

\textbf{Data collection.} We started with the manually collected and annotated dataset of 300 websites  described in \citet{SoeNGS20}. The dataset of the \citet{SoeNGS20} is available on \url{https://github.com/videoworkflow/cookiepopup}. We split  this dataset and use it in training and testing. This data set describes the cookie banners that were encountered in each of the visited websites. Each website was visited on a browser running on a laptop computer in an Incognito mode by a reviewer who recorded information about the websites containing cookie banners. 
All websites were news outlets, in English or in a Scandinavian language. 

The list of features and values from the dataset for  the cookie banner from Vice.com is listed in Table~\ref{table:vicesamplefeatures} and labels are listed in Table~\ref{table:vicesamplelabels}. 
The description of the meaning of the features follows.  
   

\begin{table}[h!]

\resizebox{\textwidth}{!}{
    \begin{tabular}{|p{2cm}|b||p{3cm}|b||p{3cm}|c|}
    \hline
     Feature name & Value & Feature name &	Value & Feature name & Value  \\
     \hline
     siteid & Vice & widetlevel & Yes, buttons & clarityofoptions & Very good: You easily understand what you can opt out from and not. You can opt out from everything possible by one click.  \\
    \hline
    country & The US & location & Middle of page, middle & iscookieusedlisted & Cookie categories and their purposes are described in an understandable way. All cookies are listed. \\
    \hline
     type & News & contentblocking & No & thirdparty & No\\
    \hline
     notyesoption & yes & optionswordscount &  559 & siteworkafter-rejectingcoookies & Yes \\
    \hline
     nameof-notyesoption & Configure Prefrences & clickstorejecttall & 2 & darkpatternisused & Yes\\
    \hline
     notyesclusters & 3 & notyesvisiblity & Immediate & areyousuremessage & No \\
    \hline
    \end{tabular}}
\caption{Sample data-point with features describing the cookie banner of Vice.com retrieved on July 2019.}
\label{table:vicesamplefeatures}
\end{table}
   
\begin{table}[h!]
\resizebox{\textwidth}{!}{
    \begin{tabular}{|p{2cm}|p{2cm}|p{2cm}|p{2cm}|p{2cm}|p{2cm}|}
    \rowcolor{Gray} Label name & Nagging &	Obstruction & 	Sneaking & Interface Interference &	Forced Action \\
    \hline
    Label value & No dark pattern &Confirmed&No dark pattern&Confirmed&Confirmed \\
    \hline
    \end{tabular}}
\caption{Dark pattern labels for Vice.com retrieved on July 2019 .}
\label{table:vicesamplelabels}
\end{table}

\noindent
\textbf{Information.} The information collected from each website (cookie banner) can be categorized into: basic website information, cookie banner related information and dark patterns. The basic website information consists of an URL, name of the website, country of origin of the website and type of the website (news or magazine).  The data about the cookie banner contains the following information: 

-- information related to 
whether a direct reject option is directly available in the interface, 
or alternatively is the user expected to interact with links to other interfaces or instructions for changing browser settings; 

-- information on whether the reject option, when available, is of the same type of widget as the accept option (e.g. whether both are buttons or links);

-- location of the pop up interface on the screen: 
up, bottom, centre;  

-- information on whether the website was 
accessible while the cookie pop up was active;


-- number of clicks required to "reject all" consent choices (whenever such possibility existed). 

The dataset also includes the privacy policies and the cookie policies as extracted text. In addition, it includes different cookie types used and explanation of the purpose of the cookie types for which the permission for which use is required. This information was not used in our experiments, but it has been parsed and translated into English using Google Translate and made available in our dataset and 
in the <GitHub link removed for review> file as a SQL lite database. 

\noindent
\textbf{Dark patterns in  cookie banners.} To evaluate the presence and type of dark patterns in the cookie banners in \citet{SoeNGS20} two reviewers had visited each website and independently recorded the presence of the five park patterns categories from \citet{gray_dark_2018}:  nagging, obstruction, sneaking, interface interference, and forced action.  The description of each of these patterns as given  by \citet{gray_dark_2018} and used for data recording in \citet{SoeNGS20}. It is given in Table~\ref{tab:dark} in the Appendix~\ref{sec:background}. 

\noindent\textbf{Features.}
The used dataset of \citet{SoeNGS20} contains many interesting and potentially relevant features for dark pattern detection. 
 The list of available features in the \citet{SoeNGS20} dataset and their data types are: 



\begin{enumerate}
    \item Site ID (siteid) - the identifying name of the website;
    \item Widget Level (widgetlevel) - the differences in design between the options of "accepting all" 
    and "rejecting all" in cookie banners (this is in the form of semi-structured text);
     \item ``Not yes'' (nameofnotyes) - the text within the first UI element (link or button) that only eventually leads to an opt out from tracking  (semi-structured text); 
    \item Location of the pop up (location)  - the location of the cookie pop up on the website (this is a description in the form of semi-structured text);
    \item Does the cookie banner disable the website (contentblocking) - whether the website is accessible and scrollable  while the pop up is active. It is yes/no (binary) data with some comments; 
    \item Words number on the option page (optionswordscount) - the number of words on the the first "layer" of the cookie banner options (ordinal data integers); 
    \item Number of clicks required for rejecting all consent (clickstorejecttall) - the number of clicks required to reject all possible cookie banners on the website (ordinal data integers);
    \item Does the website lists the purpose of the cookies (iscookieusedlisted) - whether the third party cookie list is provided or not (it is binary data with comments); 
    \item Were there any third party cookies on the website (thirdparty) declared - the number of third party categories used (semi-structured text); 
    \item Does the website work after rejecting all Cookies (iteworkafterrejectingcoookies) - whether the site works after all cookie purposes are rejected (it is binary data with some comments); 
    
\end{enumerate}
In addition to these listed features, the dataset also contains other features herewith mentioned:

\noindent
{\bf Comments from the  data collector} 
 reflecting upon the easiness, clarity and understandability regarding the 
information presented in the interface. 
As 
intentional lack of clarity can be an instrument of deception, and thus indicative of a dark pattern, we considered these reflections to be potentially relevant features that we would like to use in training a supervised model for dark pattern detection;

%
%
%

\noindent
\textbf{Type of widget options.}
Lastly, the dataset contains information regarding whether the actionable options are of the same type of widget and what that widget is. To be able to use it, we need to split this information 
into two features: equality of widget level, and type of widget. 
This process was used to create the features equality of widget level and type of widget we   used a tokenizer and stemmer.  This, and other pre-processing is described in Section~\ref{ssec:preprocessing}.

\subsection{Feature pre-processing}\label{ssec:preprocessing}

\noindent
\textbf{Cleaning.} Since manual annotated data contains inconsistencies from typos and comments intended to be read by humans, we cleaned up the data first. Cleaning up of the data was done with using Python scripts and manual correction of some typos on Microsoft Excel. Cleaning reduces the possibility of annotation mistakes degrading the performance of machine learning algorithms on our dataset. 

\noindent
\textbf{Text processing.} 
Text processing included the following steps: translation, sentiment analysis and clustering explained below.

\textit{Translation.}
Our dataset contains texts from multiple languages. 
Therefore, all non-English texts were translated into English language using Google Cloud Translate\footnote{\url{https://cloud.google.com/translate/docs/reference/libraries/v2/python}}  via its Python API.

\textit{Tokenization.} To remove the differences in language used in the original websites and hand annotated text, texts are parsed and simplified with NLTK (Natural Language Toolkit)\footnote{\url{https://www.nltk.org/}}  library in Python. In particular, we applied NLTK tookit's Tokenizer to remove extra spaces and punctuation marks and NLTK Stemmer to replace different forms of words with their root form (e.g. the root form of does, did and done is ``do''). 

\textit{Sentiment analysis.}
After translation of all the text fields 
sentiment analysis is performed using Google Cloud Language\footnote{\url{https://googleapis.dev/python/language/latest/usage.html}}. Sentiment classification was applied to  the collectors reflections in the dataset  about the quality of options in the cookie banner.
Sentiment classification returns two data points, namely, sentiment and magnitude. 
The \textit{sentiment} scores ranges from -1 (negative sentiment) to 1 (positive sentiment). The \textit{magnitude} represents the magnitude of the sentiment regardless of positive or negative sentiment. 
%


%

\textit{Clustering.} For "not yes" options text, we performed a clustering analysis to find groups in different "not yes" option texts. For that purpose, we used Universal Sentence Encoder \citep{cer_universal_nodate} to embed phrases used in "not yes" option into the vector space. The Universal Sentence Encoder  ensures that phrases that are similar or closer in meaning are embedded closer together in the vector space. It was necessary to use some form of embedding for the phrases as comparing differences in alphabets between two phrases 
does not work for finding the similarity in semantics. 

%

\medskip

\noindent
\textbf{Example of pre-processing.} 
For illustration purposes, we convey an example on clustering of the ``not yes'' option text feature. During the data collection, the text that indicates an \emph{alternative} to 
accepting consent 
was collected as  a separate text feature regardless of whether the text appeared on a button or as link. 

\emph{Translation.}
The ``not yes'' option text in the dataset is  unstructured and contains words in multiple languages. To be able to use it for a training
machine learning model, we firstly need to translate it into English  (as the majority of the data points were in English), which  was performed using the  Google Cloud Translate API with Python client library \footnote{\url{https://cloud.google.com/translate/docs/reference/libraries/v3/python}}. Each ``not yes'' option text was individually fed into the API to ensure that each of them is processed on its own.
%
%

\emph{Clustering.} The translated text was inspected manually, and we observed that the phrasing of the text label ``not yes'' option varies.
A number of different 
but similar phrases, e.g. \emph{``Read more''} and \emph{``More information''}. 
 As per Figure~\ref{fig:cookieconsent}, it would be "Configure Preferences", though other  
options ranges from \emph{``Learn More''} to \emph{``Options''} are found in the dataset. 
Therefore, clustering is applied to discover different categories or clusters of the ``not yes'' text.  First, the  ``not yes'' phrases, now all translated to English, were encoded into vectors using  the Universal Sentence Encoder \citep{cer_universal_nodate}. Text embedding -- a popular method used in Natural Language Processing before tasks such as clustering, translating, classification and similarity --, in our case with Universal Sentence Encoder, converts the phrases 
into  512 dimensional vectors or an array with 512 values.

\begin{figure}[h]
    \includegraphics[width=0.8\linewidth]{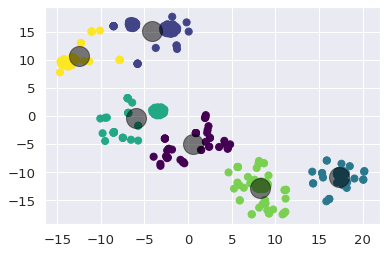}
    \caption{Clustering of text of ``not yes'' option with PCA visualization: small circles filled with six different colors correspond to different ``not yes'' text data points. The color represents membership to a cluster with the same color. 
The center of each clusters is marked with larger circles in gray color. 
The x-axis and y-axis are two  dimensions computed by PCA from a combination of the initial 512 dimensions.}
    \label{fig:clustering}
\end{figure}

\emph{Visualization.}
Since it is impossible to visualize 512 dimensions, we used  Principle Component Analysis (PCA) ~\citep{PCA} to reduce the dimensions to 2 so that the text embedding can be plotted and visually analyzed.  
It was quite clear from the visualization that the data can be clustered into six clusters. We used K-means method \citep{Lloyd82leastsquares} for clustering. 
The clusters are visualized and the resulting clusters are plotted in Figure \ref{fig:clustering}.

\vspace{2mm} 

The rest of the features which are not mentioned in this subsection only required general clean up such as removal of special characters, correction of typos and removal of additional comments from the reviewer. 
%
%
The final set of cleaned and transformed features used to train a prediction model for dark pattern detection is summarized in the Table~\ref{table:features}. 

\begin{table}[h!]
 \resizebox{\textwidth}{!}{
\begin{tabular}{|p{3cm}|p{4cm}||p{2cm}|p{4cm}|} 
\hline
 Feature & Description & Type & Example values \\
\hhline{====}
notyesclusters & The text on the not yes button or link after assigning clusters & categorical & six clusters from 0 to 5  \\ 
\hline
equalwidgetlevel & Whether the accept and not yes are of the same widget level (button and button) & binary & Yes/No  \\  
\hline
widgettypelevel & The type of the not yes widget & categorical & button, Link, box, drop-down  \\ 
\hline
location & Location of the popup on the website & categorical & Middle of page, bottom entire, top entire  \\  
\hline
contentblocking & Whether website is accessible when the pop up active  & binary & Yes/No  \\  
\hline
optionswordscounted & Words on the options page counted.  & integer &   \\  
\hline
clickstorejectall & Number of clicks required to reject all third party consents& integer &  \\  
\hline
notyesvisibility & The visibility of the not yes option. & categorical & immediate, scroll  \\  
\hline
clarityofoptions & Sentiment value of the clarity of option comment  & float & -1 to 1  \\  
\hline
iscookieusedlisted & Sentiment value for whether the third party cookie used is listed clearly & float &  -1 to 1 \\  
\hline
\end{tabular}}
\caption{Final feature set for dark pattern classification, their description, type and example values.}
\label{table:features}
\end{table}

\subsection{Data labeling}
The dataset of Soe et al.~\citep{SoeNGS20} 
was not 
collected with the purpose of being used as training data in machine learning. 
The data points in that dataset were ``labeled" with information on the identified types of dark patterns from  \citep{gray_dark_2018}. 
However, the identification of these dark patterns was done by two independent reviewers for each data-point, who acted without coordinating or agreeing on how to identify the patterns. 
The reviewers more often than not disagreed on which specific dark pattern was present.
As a result,  there were a lot of inconclusive results among the reviewers and also some disagreement of whether there was indeed a dark pattern (one reviewer noted the presence of dark pattern, while it was not noted by the other). 

We needed to find a way to work with these somewhat ambiguous labels. 
We adopted the following  labelling:  
\begin{itemize}
    \item \emph{No dark pattern usage detected}  (Integer value 0) is assigned when both reviewers noted the absence of a dark pattern. No dark pattern usage found by both reviewers.  
    \item \emph{Possible dark pattern usage} (Integer value 1) is assigned when only one reviewer noted the presence of a dark pattern. Possible dark pattern usage. 
    \item \emph{Confirmed dark pattern usage} (Integer value 2)  is assigned when both reviewers noted the presence of a dark pattern. Dark pattern usage is confirmed by both reviewers.
\end{itemize}

\vspace{1mm} 

In this work, we 
train the 
machine learning model with these three labels, thereby  classifying the dataset into: 
samples with no dark pattern usage, 
samples with 
possible dark pattern usage (only one reviewer found it), and 
samples with confirmed dark pattern usage. 

The distribution of the labels is depicted in  Table~\ref{table:numlabels}. Therein, the labels are not evenly distributed, and in particular,  nagging and sneaking dark pattern categories have negligible or no confirmed samples. 
This is the result of the difficulties in detecting these dark patterns mentioned by the collectors of the dataset \citep{SoeNGS20}.

\begin{table}[h!]
\begin{tabular}{|p{4cm}|p{2.5cm}|p{2.5cm}|p{2.5cm}|} 
 \hline
 Dark Pattern & No dark pattern - 0 & Possible - 1 & Confirmed - 2\\
\hhline{====}
Nagging & 229 & 68 & 3  \\ 
\hline
Obstruction & 50 & 121 & 129  \\ 
\hline
Sneaking & 186 & 114 & 0 \\ 
\hline
Interface Interference & 55 & 109 & 136 \\
 \hline
Forced action & 181 & 88 & 31 \\
 \hline
\end{tabular}
\caption{Number of labels for five dark patterns}
\label{table:numlabels}
\end{table}

\subsection{Feature importance measure}
The features and labels described are used for feature importance estimation by using the Random Forest Classification\footnote{\url{https://scikit-learn.org/stable/modules/generated/sklearn.ensemble.RandomForestClassifier.html}} model in scikit learn. 
We used this feature importance measure to help us understand and interpret features and labels, but we did not use this for training machine learning classifiers. The obtained results are depicted in Figures~\ref{fig:sub-first},~\ref{fig:sub-second},~\ref{fig:sub-third},~\ref{fig:sub-fourth},  and~\ref{fig:sub-figth}.  The feature is given in the Y-axis and the feature importance, measured on a scale of 0 to 1, is given on the X-axis.

We observe that not all of the feature importance values were equally useful in interpreting the data. 
For example, \emph{location} of the pop up as a feature is estimated to be highly important for identifying the dark pattern nagging, see  Figure~\ref{fig:sub-first}, which makes sense and already confirmed in the study of \citet{utz_informed_2019}, as a pop up that is displaying in the middle of the page 
can be considered as nagging every time a user browses the web. 
However, the two features estimated as most important for the dark pattern forced action -- the \textit{purposes}  (\emph{iscokieusedlised}) and \textit{clarity of options} (\emph{clarityofoptions}), see Figure~\ref{fig:sub-figth}, do  not really make sense.

\section{Classification of dark patterns using machine learning}\label{sec:classification}

The type of features and number of data points in the dataset very much dictated the choice of supervised machine learning classifier we could use. We chose to use the Gradient Boosted Tree method for classifying the detected dark patterns\footnote{\url{https://scikit-learn.org/stable/modules/generated/sklearn.ensemble.GradientBoostingClassifier.html}} as a mix of categorical and continuous features is best solved by decision tree-based classification \citep{friedman_greedy_2001}. Gradient Boosted Tree \citep{friedman_greedy_2001} for classification uses a combination of many small decision trees in which each small decision trees tries to improve on the results of the combination of previously built trees. Decision tree is a method using branches on each feature to divide the dataset into smaller subsets that contain more homogeneous samples. 

The training process involves two steps: tuning the hyper-parameters for the classifier and then training the classifier. The hyper-parameter tuning, finding optimal parameters for Gradient Boosted Tree with our dataset, is done with GridSearch\footnote{\url{https://scikit-learn.org/stable/modules/generated/sklearn.model_selection.GridSearchCV.html}} and the following parameters are used learning rate: (0.15, 0.1, 0.05, 0.01, 0.005, 0.001) and n\_estimators: (10, 15, 20, 25, 30, 35, 40). This resulted in optimal parameters for our training which are {\emph{learning\_rate: 0.01} and \emph{n\_estimators: 30}}. 


The dataset is split into training dataset, for which we used two thirds of our dataset, and testing dataset, for which we used one third. The split was done by randomly assigning data points to either the test or training data set. 

Five different ``Gradient Boosted Tree classifiers'' are trained, one for each of the five dark patterns. The models are trained independently for each dark pattern and each of the five models attempts to predict whether a presented dark pattern is of three class labels, namely, 0-No dark pattern usage detected, 1-Possible dark pattern usage and 2- Confirmed dark pattern usage. Table~\ref{table:accuracy_score} list the accuracy score the machine learning models for each type of dark patterns categories. The accuracy score is the mean accuracy score for each of the three class labels in our test dataset weighted by their numbers relative to the total samples in the test dataset. For each of the class label the accuracy score is computed as \emph{number of correctly identified samples} divided by \emph{total numbers of samples with that label} in the test dataset. As we can observe, the worst accuracy is obtained for the Interface Interference dark pattern, just 0.535,  which is a still better than random -- a completely random classifier will achieve 0.33 accuracy in this case. 

\begin{table}[h!]
\begin{tabular}{|p{6cm}|p{6cm}|} 
 \hline
 Dark Pattern & Accuracy score \\
\hhline{==}
Nagging & 0.720  \\ 
\hline
Obstruction & 0.500  \\ 
\hline
Sneaking & 0.686  \\ 
\hline
Interface Interference & 0.570 \\
 \hline
Forced action & 0.628 \\
 \hline
\end{tabular}
\caption{Accuracy scores of dark pattern recognition}
\label{table:accuracy_score}
\end{table}

Since we are dealing  with  a  multi-class classification problem, confusion matrices were created for each of the classifiers. These confusion matrices add to the information provided in the accuracy table by displaying the performance in terms of predicted labels and actual labels. The rows are actual labels in our dataset and columns are predicted labels from our machine learning models. The cells represent the ratio of actual label that are classified as predicted labels for the corresponding rows and columns. The diagonal line from top left to bottom right represents the values of correct predictions and the rest of the cells are incorrect prediction of different kinds. These matrices are given in Figure~ \ref{fig:confusion} in the Appendix~\ref{APPENDIX}.

\section{Discussion}\label{sec:discussion}

Though the raw numbers from this report are not  encouraging, this work reveals a lot for the path towards automatically detection of deceptive UI design in cookie banners. In this section we discuss the most prevalent detected dark patterns type, lessons learned and policy implications. 

\medskip

\noindent
\textbf{Prevalence of dark pattern type.} During the exercise, it is obvious that \emph{Forced action} is one of the easiest dark pattern to identify for the reviewers. Consequently, it has the most accurate labels out of all the dark patterns.
The accuracy results and confusion matrix scored on the trained classifier for different dark patterns also showed that it is best as detecting this type of dark pattern of forced action. In contrast, nagging and sneaking are most difficult to automatically detect, which is not surprising given the very low number of examples of confirmed presence of these patterns in the dataset. 
%
For the rest of the dark patterns, the automated classification is plagued by difficulties for reviewers in identifying the dark patterns, as explained below. Further work would be needed regarding implicit characteristics on the other categories.

\medskip

\noindent
\textbf{Lessons learned.} From our experiment and the yielding difficulties observed, we would like to share the most important takeaways that  can inform other automated dark pattern detection initiatives based on ML.

1- \textit{Labelling dataset and codebook.} Any successful automated ML detection  depends on tuning dark patterns classifications with concrete features of cookie banners.
And an automated approach to assess  cookie banners is surely difficult  in practice due to their differing
designs~\citep{Sanchez-Rola-cookies2019,Degeling2019WeVY,Kretschmer-cookiebanners21}. This fact entails that the analysis of cookie banners usually introduces a significant amount
of manual labeling effort~\citep{Sanchez-Rola-cookies2019,Kretschmer-cookiebanners21}.
In fact, the current dataset from \cite{SoeNGS20} is not yet suitable for automatic analysis. Better labels are necessary for the dataset and it can be achieved by  clarifying the identification 
of dark patterns by reviewers. Doing so requires a dark pattern classification which is different from \citet{gray_dark_2018} and thus more specific and amenable to  cookie banners. 
Accordingly, additional research is needed to improve codebook consensus across dark pattern classification and its inherent characteristics, and alongside accounting for newly identified patterns in future work~\citep{Gunawan-comparativestudy2021}. On this stance,  \citet{gray-legalreq} propose for an holistic, n-dimensional dark patterns analysis in furtherance for such a consensual dark patterns definition.

2- \textit{Guidelines.} Such codebook could be further coupled with guidelines for reviewers quality labels. 
The guidelines should be specific enough to allow for maximal agreement among reviewers on which dark pattern is present. This in turn imposes again the request for a better definition of dark patters. There is a limit to how precisely a dark pattern can be defined since when doing so, one can also abolish it. Dark patterns are tricky because ``It is rarely possible to foresee which new
patterns are going to emerge, and as a result, detection measures are always reactive, and rely on practitioners that
constantly update the existing pattern databases as well as engaged consumers that point out new occurrences''.  \citep{Hausner2021}. Thus, guidelines may need to be continuously updated and limited to a particular context of use. 

3 - \textit{Clustering.} The process for redefining the dark pattern category for cookie banners could also be done using machine learning by considering a corpus of cookie banners and analyzing clusters within them. Similarity patterns among cookie banners might yield new insights to what new dark pattern definitions can be. We can then use these patterns which are more ``visible'' to a machine to train a prediction model. 

4 - \textit{Mixed approach.} An automated approach combined with manual methods could ensure better results towards detecting dark patterns in cookie banners. As we discussed in the Introduction, some features can be difficult to harvest by machines but can be precisely defined and easily identified by people, even when the people do not necessarily agree on which dark pattern they are looking at. Some dark patterns, such as nagging, would only be detectable with a human observing their behaviour.   %

5 - \textit{Dark Patterns conceptual refinement.} Need for a refinement of the concept of dark patterns -- per dark pattern category (following the cognition of  \citet{Mathur-2021,geronimo}. As \citet{Gunawan-comparativestudy2021} posits, additional research is needed to develop the theory of dark pattern-blindness and
potential mitigation strategies in order to detect more accurately the presence of dark patterns.


\medskip 

\noindent
\textbf{Policy and legal implications.}
Any  AI computational system  aiming to detect dark patterns should align to detectable issues that are already deemed illegal by authoritative sources. 
But there are only \textit{few} (mandatory) legal rules in Europe constraining the use of dark patterns, and enforcement is slow in holding websites accountable.
The only mandatory decision ascertaining any UI based aspect is dated of 2019 forbidding the use of pre-ticked boxes~\citep{CJEU-Planet49-19}. 
From Article 7(3) of the GDPR (\textit{"it shall be as easy to withdraw as to give consent"}, it can be interpreted that privacy choices should be equal (e.g. parity in accept, reject and revoke choices)~\citep{Sant-etal-20-TechReg, nouwens_dark_2020}. Parity feature entails i) equal widgets, ii) equal number of times to either accept/reject/revoke consent, iii) across modalities (web, mobile and app setting levels)~\citep{Gunawan-comparativestudy2021}. This reasoning on feature parity needs still to be held definitive by court decisions as well for consistency in all EU.
The ePrivacy Regulation draft\footnote{https://data.consilium.europa.eu/doc/document/ST-6087-2021-INIT/en/pdf}, being discussed in the European Council, as of today, is absent on the definition of dark patterns or UI features, even accepting the use of cookie walls (Council's version), considered as an onstructive dark pattern~\citep{Kretschmer-cookiebanners21,gray-legalreq}.
Such weak enforcement and the high rate of consent optimization enhanced by using faulty designs in cookie banners~\citep{hils-2020, Santos-cmp21} at scale, facilitated by the use of consent management platforms, explain the recurrent use of dark patterns in cookie banners.
In the future, we need to see a more serious approach to enforcement, either by courts, or by decisions issued by data protection authorities. That is the only way to ensure that automated systems can rely on the necessary legal certainty in identifying dark patterns by identifying concrete characteristics of design.   



\section{Related Work}\label{sec:relatedwork}

Detecting and quantifying the presence of dark patterns  has deserved vibrant attention, concretely, within privacy policies~\citep{sleightsprivacy-adjerid}, e-commerce websites~\citep{Mathur-2021}, popular mobile apps~\citep{nouwens_dark_2020, Sanchez-Rola-cookies2019, utz_informed_2019,geronimo}, video games~\citep{Zagal2013DarkPI},  cookie banners, among other contexts.
However,  there is little work reporting use of machine learning for  automated detection of dark patterns in user interfaces of cookie banners. In this section we analyse automated detection of dark patterns in general and in cookie banners.

\medskip

\noindent
\textbf{Detection of dark patterns in online services.}
 \citet{mathur_dark_2019} analyzed $\sim$53K product pages from $\sim$11K shopping websites, and 
discovered 1,818 dark pattern instances, together representing 15 types and 7 broader categories. 
The goal of their work is to present ``automated techniques that enable
experts to identify dark patterns on a large set of websites in one particular category''. 
 The process presented is done in three steps:. The first step is corpus creation through crawling Alexa top websites which are ranked according to monthly web traffic. From that list of top traffic websites, a tool called Webshrinker is used to categorize into shopping or not shopping categories. After that, only English language websites are kept resulting in the final list of 19,455 shopping websites. 
 The second step is data collection of the product checkout pages. The third step is data analysis on the product checkout pages using  using Hierarchical Density-based Spatial Clustering of Applications with Noise (HDBSCAN) \citep{hutchison_density-based_2013}. This is an unsupervised learning method that can form hierarchical clusters from the dataset. Samples from these clusters are then manually examined to identify different categories of dark patterns in product checkout pages.

 \citet{CurleyOGTS2021}
developed a framework for 
automated detection of potential instances of web-based dark 
patterns. 
They  identify whether or not it is 
technically possible to automatically detect that particular 
pattern.
They analyze known dark patterns in terms of 
whether they can be: 
(1) detected in an automated way (either 
partially or fully), 
(2) detected in a manual way (either 
partially or fully) and
(3) cannot be detected at all due to  variation in either how the 
pattern is defined or implemented, 
there is no direct way of detecting which hampers web 
crawling and web scraping techniques.
 Some patterns are easier to detect than others, and 
 some are impossible to detect in an automated 
fashion. 
They propose a software tool
that can automatically alert users 
of the presence of web-based dark patterns

\medskip

\noindent
\textbf{Detection of dark patterns in cookie banners.} 
We focus on related where where some attempt for automatic detection of dark patterns was presented. 
 %
%
 \citet{nouwens_dark_2020} performed a study on the five most popular
CMPs on the top 10,000 websites in the UK which has yielded 680
banners. They aimed to study the impact of
various designs of consent banners, user interface design nudges
and level of granularity of options. They used a hand-crafted approach for dark patterns detection. The implementation details for  using hand-crafted scripts for detection is not available 
 and it only works just for 6.8\% of the 10,000 websites crawled. 
 The authors looked at  unambiguous, widget level (easiness to reject), 
presence of pre-ticked boxes.
 \citet{matte_cookie_2020} used semi-automatic methods and only made the content of the cookie banner notifications available to the users via a script and human labour is used to identify four GPDR violations, as depicted in Table~\ref{table:gdpr_conditions_studies}:
 consent was stored before the user made
the choice,whether a cookie banner offers a way to opt out,  whether
there were pre-selected choices, if the choice that the
user had made was respected at all.
 %
  %
 %
 %
 \citet{Hausner2021} presented ongoing work in the direction of automatic
detection of dark patterns on cookie banners. They use the feature representation for cookie banners and automatically extract the feature values.   Their goal is to build a general
framework to detect dark patterns on arbitrary web pages (regardless of their domain). Unlike us, they do not consider dark patters as abstract concepts, but  focus on widget parity.
Their approach does not consider dynamic aspects of dark patterns.
By applying the implemented algorithm to more than 4000 German websites
extracted from a list of the top one
million web sites according to Alexa.com, around 2800 cookie banners were extracted and analyzed. By utilizing
features like the HTML tag of elements, they  obtained a large amount of clickable elements within the
banners. They extracted textual features from those elements, and used clustering techniques to find different groups
of buttons with regard to their textual content. Based on the initial clustering and a manual relabeling of critical items,
a Support Vector Classifier is trained to distinguish between multiple button types. 

 %
 %
 The GDPR violations features 
 detected in two of these works in available in Table~\ref{table:gdpr_conditions_studies} in the Appendix~\ref{APPENDIX}.

\section{Conclusion}\label{sec:conclusion}

We considered the problem of automatically detecting the five dark patterns of \cite{gray_dark_2018} by training a supervised machine learning model using the data set of \cite{SoeNGS20}. Our approach can be considered naive, since the data set used was relatively small and not well balanced - dark patterns of different type occur with different frequency in the data set. However, as we discussed throughout the paper, creating a training dataset is a considerable challenge in its own right. It requires a combination of human effort, automated extraction of feature values and pre-processing of extracted material using various AI technologies such as clustering and sentiment analysis.  
Furthermore, unlike related work, we attempt to detect dark patterns directly, rather than specific features that make the pattern dark, like for example widget inequality, which is much less ambitious. 

We used this experiment primarily as a spring-board to better understand the problems of using machine learning for automating dark pattern detection. We present a detailed analysis on the challenges involved and consider this our main contribution. Our experiment allows us to clearly outline promising directions of future work. 

Clearly, a pipeline of various methods and techniques would eventually need to be constructed for a functioning automated dark pattern detection tool. To construct it we need to decompose the experience:  dark patterns into elements that can be automatically, or at least easily processed. This is a task that requires expertise in human-computer interaction and cognitive science. 
It is also a  task that is perhaps easiest tackled when focusing on one specific domain at a time. For example cookie banners should be considered as one domain, whereas privacy settings another.

Visual cues and components of a dark pattern should be explored as an image recognition problem. To this end, we need to construct a common resource, a collection of labeled images, one for each visual cue. For example: one collection of images that contain screenshots of cookie banners with different level widgets. 

What the \citet{SoeNGS20} dataset shows,  is that there is a lot of variety in the textual information that describes the purposes of the data collection, cookies, trackers etc. We did not explore how to use this text in our approach, but one should consider that the dark pattern here is the volume and language style of the text, rather than the information it is supposed to convey. As with images, a common resource needs to be created to create training data for a supervised learning algorithm that would label text as confusing or comprehensible, using feature that describe the volume of text and various linguistic cues such as legalese. 

In our work, we were not able to identify clear distinguishable features that would  discern among the dark patterns. This is because we worked with a ``found'' data set. Namely, the features were not engineered to represent a specific dark pattern. 

%
%
%
\citet{SoeNGS20} also suggested 12 refined dark patterns but did not provide a label for these dark patterns in their original dataset. An immediate first step would be to creation a new dataset with similar categories of dark patterns as proposed in \citep{SoeNGS20} and features engineered to discern among those dark patterns. 


\typeout{} 
\bibliographystyle{ACM-Reference-Format}
\bibliography{biblio}


\begin{thebibliography}{67}


\ifx \showCODEN    \undefined \def \showCODEN     #1{\unskip}     \fi
\ifx \showDOI      \undefined \def \showDOI       #1{#1}\fi
\ifx \showISBNx    \undefined \def \showISBNx     #1{\unskip}     \fi
\ifx \showISBNxiii \undefined \def \showISBNxiii  #1{\unskip}     \fi
\ifx \showISSN     \undefined \def \showISSN      #1{\unskip}     \fi
\ifx \showLCCN     \undefined \def \showLCCN      #1{\unskip}     \fi
\ifx \shownote     \undefined \def \shownote      #1{#1}          \fi
\ifx \showarticletitle \undefined \def \showarticletitle #1{#1}   \fi
\ifx \showURL      \undefined \def \showURL       {\relax}        \fi
\providecommand\bibfield[2]{#2}
\providecommand\bibinfo[2]{#2}
\providecommand\natexlab[1]{#1}
\providecommand\showeprint[2][]{arXiv:#2}

\bibitem[\protect\citeauthoryear{AccessNow}{AccessNow}{2020}]%
        {AccessNow-2020}
\bibfield{author}{\bibinfo{person}{AccessNow}.}
  \bibinfo{year}{2020}\natexlab{}.
\newblock \bibinfo{title}{Two years under the EU GDPR. State of play, analysis
  and recommendations. An implementation progress report. Two years under the
  EU GDPR.}
\newblock
\newblock
\urldef\tempurl%
\url{https://www.accessnow.org/cms/assets/uploads/2020/05/Two-Years-Under-GDPR.pdf}
\showURL{%
\tempurl}


\bibitem[\protect\citeauthoryear{Acquisti}{Acquisti}{2009}]%
        {Acquisti-nudging}
\bibfield{author}{\bibinfo{person}{Alessandro Acquisti}.}
  \bibinfo{year}{2009}\natexlab{}.
\newblock \showarticletitle{Nudging Privacy: The Behavioral Economics of
  Personal Information}.
\newblock \bibinfo{journal}{\emph{IEEE Security Privacy}} \bibinfo{volume}{7},
  \bibinfo{number}{6} (\bibinfo{year}{2009}), \bibinfo{pages}{82--85}.
\newblock
\urldef\tempurl%
\url{https://doi.org/10.1109/MSP.2009.163}
\showDOI{\tempurl}


\bibitem[\protect\citeauthoryear{Adjerid, Acquisti, Brandimarte, and
  Loewenstein}{Adjerid et~al\mbox{.}}{2013}]%
        {sleightsprivacy-adjerid}
\bibfield{author}{\bibinfo{person}{Idris Adjerid}, \bibinfo{person}{Alessandro
  Acquisti}, \bibinfo{person}{Laura Brandimarte}, {and} \bibinfo{person}{George
  Loewenstein}.} \bibinfo{year}{2013}\natexlab{}.
\newblock \showarticletitle{Sleights of Privacy: Framing, Disclosures, and the
  Limits of Transparency}. In \bibinfo{booktitle}{\emph{Proceedings of the
  Ninth Symposium on Usable Privacy and Security}} (Newcastle, United Kingdom)
  \emph{(\bibinfo{series}{SOUPS '13})}. \bibinfo{publisher}{Association for
  Computing Machinery}, \bibinfo{address}{New York, NY, USA}, Article
  \bibinfo{articleno}{9}, \bibinfo{numpages}{11}~pages.
\newblock
\showISBNx{9781450323192}
\urldef\tempurl%
\url{https://doi.org/10.1145/2501604.2501613}
\showDOI{\tempurl}


\bibitem[\protect\citeauthoryear{{Article 29 Working Party}}{{Article 29
  Working Party}}{2013}]%
        {WP208-guidance-cookies}
\bibfield{author}{\bibinfo{person}{{Article 29 Working Party}}.}
  \bibinfo{year}{2013}\natexlab{}.
\newblock \bibinfo{title}{{ Working Document 02/2013 providing guidance on
  obtaining consent for cookies’ (WP208, 2 October 2013)}}.
\newblock
\newblock


\bibitem[\protect\citeauthoryear{Auxier, Rainie, Anderson, Perrin, Kumar, and
  Turner}{Auxier et~al\mbox{.}}{2019}]%
        {Auxier2019}
\bibfield{author}{\bibinfo{person}{Brooke Auxier}, \bibinfo{person}{Lee
  Rainie}, \bibinfo{person}{Monica Anderson}, \bibinfo{person}{Andrew Perrin},
  \bibinfo{person}{Madhu Kumar}, {and} \bibinfo{person}{Erica Turner}.}
  \bibinfo{year}{2019}\natexlab{}.
\newblock \bibinfo{booktitle}{\emph{{Americans and Privacy: Concerned, Confused
  and Feeling Lack of Control Over Their Personal Information}}}.
\newblock \bibinfo{type}{{T}echnical {R}eport}. \bibinfo{institution}{Pew
  Research Center}. \bibinfo{pages}{1--63} pages.
\newblock


\bibitem[\protect\citeauthoryear{Bauer, Bergstrøm, and Foss-Madsen}{Bauer
  et~al\mbox{.}}{2021}]%
        {bauer_are_2021}
\bibfield{author}{\bibinfo{person}{Jan~M. Bauer}, \bibinfo{person}{Regitze
  Bergstrøm}, {and} \bibinfo{person}{Rune Foss-Madsen}.}
  \bibinfo{year}{2021}\natexlab{}.
\newblock \showarticletitle{Are you sure, you want a cookie? – The effects of
  choice architecture on users’ decisions about sharing private online data}.
\newblock \bibinfo{journal}{\emph{Computers in Human Behavior}}
  (\bibinfo{year}{2021}), \bibinfo{pages}{106729}.
\newblock
\showISSN{07475632}
\urldef\tempurl%
\url{https://doi.org/10.1016/j.chb.2021.106729}
\showDOI{\tempurl}


\bibitem[\protect\citeauthoryear{Bengio, LeCun, and Hinton}{Bengio
  et~al\mbox{.}}{2021}]%
        {BengioLH21}
\bibfield{author}{\bibinfo{person}{Yoshua Bengio}, \bibinfo{person}{Yann
  LeCun}, {and} \bibinfo{person}{Geoffrey~E. Hinton}.}
  \bibinfo{year}{2021}\natexlab{}.
\newblock \showarticletitle{Deep learning for {AI}}.
\newblock \bibinfo{journal}{\emph{Commun. {ACM}}} \bibinfo{volume}{64},
  \bibinfo{number}{7} (\bibinfo{year}{2021}), \bibinfo{pages}{58--65}.
\newblock
\urldef\tempurl%
\url{https://doi.org/10.1145/3448250}
\showDOI{\tempurl}


\bibitem[\protect\citeauthoryear{Bhoot, Shinde, and Mishra}{Bhoot
  et~al\mbox{.}}{2020}]%
        {Bhoot2020TowardsTI}
\bibfield{author}{\bibinfo{person}{Aditi~M. Bhoot}, \bibinfo{person}{Mayuri~A.
  Shinde}, {and} \bibinfo{person}{Wricha~P. Mishra}.}
  \bibinfo{year}{2020}\natexlab{}.
\newblock \showarticletitle{Towards the Identification of Dark Patterns: An
  Analysis Based on End-User Reactions}.
\newblock \bibinfo{journal}{\emph{IndiaHCI '20: Proceedings of the 11th Indian
  Conference on Human-Computer Interaction}} (\bibinfo{year}{2020}).
\newblock


\bibitem[\protect\citeauthoryear{Bongard-Blanchy, Rossi, Rivas, Doublet,
  Koenig, and Lenzini}{Bongard-Blanchy et~al\mbox{.}}{2021}]%
        {Bongard2021manipulated}
\bibfield{author}{\bibinfo{person}{Kerstin Bongard-Blanchy},
  \bibinfo{person}{Arianna Rossi}, \bibinfo{person}{Salvador Rivas},
  \bibinfo{person}{Sophie Doublet}, \bibinfo{person}{Vincent Koenig}, {and}
  \bibinfo{person}{Gabriele Lenzini}.} \bibinfo{year}{2021}\natexlab{}.
\newblock \showarticletitle{“I am definitely manipulated, even when I am
  aware of it. It’s ridiculous!” - Dark Patterns from the End-User
  Perspective}.
\newblock \bibinfo{journal}{\emph{Proceedings of ACM DIS Conference on
  Designing Interactive Systems}} (\bibinfo{year}{2021}).
\newblock
\urldef\tempurl%
\url{https://doi.org/10.1145/3461778.3462086}
\showDOI{\tempurl}


\bibitem[\protect\citeauthoryear{B\"{o}sch, Erb, Kargl, Kopp, and
  Pfattheicher}{B\"{o}sch et~al\mbox{.}}{2016}]%
        {Bosch2016}
\bibfield{author}{\bibinfo{person}{Christoph B\"{o}sch},
  \bibinfo{person}{Benjamin Erb}, \bibinfo{person}{Frank Kargl},
  \bibinfo{person}{Henning Kopp}, {and} \bibinfo{person}{Stefan Pfattheicher}.}
  \bibinfo{year}{2016}\natexlab{}.
\newblock \showarticletitle{Tales from the Dark Side: Privacy Dark Strategies
  and Privacy Dark Patterns}.
\newblock \bibinfo{journal}{\emph{Proceedings on Privacy Enhancing
  Technologies}} \bibinfo{volume}{2016}, \bibinfo{number}{4}
  (\bibinfo{year}{2016}), \bibinfo{pages}{237--254}.
\newblock
\urldef\tempurl%
\url{https://doi.org/10.1515/popets-2016-0038}
\showDOI{\tempurl}


\bibitem[\protect\citeauthoryear{Brave}{Brave}{2020}]%
        {brave-2020}
\bibfield{author}{\bibinfo{person}{Brave}.} \bibinfo{year}{2020}\natexlab{}.
\newblock \bibinfo{title}{Europe’s governments are failing the GDPR.
  Brave’s 2020 report on the enforcement capacity of data protection
  authorities.}
\newblock
\newblock
\urldef\tempurl%
\url{https://brave.com/wp-content/uploads/2020/04/Brave-2020-DPA-Report.pdf}
\showURL{%
\tempurl}


\bibitem[\protect\citeauthoryear{Brignull, Miquel, Rosenberg, and
  Offer}{Brignull et~al\mbox{.}}{2015}]%
        {brignull_dark_2015}
\bibfield{author}{\bibinfo{person}{Harry Brignull}, \bibinfo{person}{Marc
  Miquel}, \bibinfo{person}{Jeremy Rosenberg}, {and} \bibinfo{person}{James
  Offer}.} \bibinfo{year}{2015}\natexlab{}.
\newblock \bibinfo{title}{Dark {Patterns} - {User} {Interfaces} {Designed} to
  {Trick} {People}}.
\newblock
\newblock
\urldef\tempurl%
\url{https://www.darkpatterns.org/}
\showURL{%
\tempurl}
\newblock
\shownote{Library Catalog: www.darkpatterns.org.}


\bibitem[\protect\citeauthoryear{Campello, Moulavi, and Sander}{Campello
  et~al\mbox{.}}{2013}]%
        {hutchison_density-based_2013}
\bibfield{author}{\bibinfo{person}{Ricardo J. G.~B. Campello},
  \bibinfo{person}{Davoud Moulavi}, {and} \bibinfo{person}{Joerg Sander}.}
  \bibinfo{year}{2013}\natexlab{}.
\newblock \showarticletitle{Density-{Based} {Clustering} {Based} on
  {Hierarchical} {Density} {Estimates}}. In \bibinfo{booktitle}{\emph{Advances
  in {Knowledge} {Discovery} and {Data} {Mining}}},
  \bibfield{editor}{\bibinfo{person}{David Hutchison}, \bibinfo{person}{Takeo
  Kanade}, \bibinfo{person}{Josef Kittler}, \bibinfo{person}{Jon~M. Kleinberg},
  \bibinfo{person}{Friedemann Mattern}, \bibinfo{person}{John~C. Mitchell},
  \bibinfo{person}{Moni Naor}, \bibinfo{person}{Oscar Nierstrasz},
  \bibinfo{person}{C.~Pandu~Rangan}, \bibinfo{person}{Bernhard Steffen},
  \bibinfo{person}{Madhu Sudan}, \bibinfo{person}{Demetri Terzopoulos},
  \bibinfo{person}{Doug Tygar}, \bibinfo{person}{Moshe~Y. Vardi},
  \bibinfo{person}{Gerhard Weikum}, \bibinfo{person}{Jian Pei},
  \bibinfo{person}{Vincent~S. Tseng}, \bibinfo{person}{Longbing Cao},
  \bibinfo{person}{Hiroshi Motoda}, {and} \bibinfo{person}{Guandong Xu}}
  (Eds.), Vol.~\bibinfo{volume}{7819}. \bibinfo{publisher}{Springer Berlin
  Heidelberg}, \bibinfo{address}{Berlin, Heidelberg},
  \bibinfo{pages}{160--172}.
\newblock
\showISBNx{978-3-642-37455-5 978-3-642-37456-2}
\urldef\tempurl%
\url{https://doi.org/10.1007/978-3-642-37456-2_14}
\showDOI{\tempurl}
\newblock
\shownote{Series Title: Lecture Notes in Computer Science.}


\bibitem[\protect\citeauthoryear{Cer, Yang, yi~Kong, Hua, Limtiaco, John,
  Constant, Guajardo-Cespedes, Yuan, Tar, Sung, Strope, and Kurzweil}{Cer
  et~al\mbox{.}}{2018}]%
        {cer_universal_nodate}
\bibfield{author}{\bibinfo{person}{Daniel Cer}, \bibinfo{person}{Yinfei Yang},
  \bibinfo{person}{Sheng yi Kong}, \bibinfo{person}{Nan Hua},
  \bibinfo{person}{Nicole Limtiaco}, \bibinfo{person}{Rhomni~St. John},
  \bibinfo{person}{Noah Constant}, \bibinfo{person}{Mario Guajardo-Cespedes},
  \bibinfo{person}{Steve Yuan}, \bibinfo{person}{Chris Tar},
  \bibinfo{person}{Yun-Hsuan Sung}, \bibinfo{person}{Brian Strope}, {and}
  \bibinfo{person}{Ray Kurzweil}.} \bibinfo{year}{2018}\natexlab{}.
\newblock \bibinfo{title}{Universal Sentence Encoder}.
\newblock
\newblock
\showeprint[arxiv]{1803.11175}~[cs.CL]


\bibitem[\protect\citeauthoryear{Chatellier, Delcroix, Hary, and
  Girard-Chanudet}{Chatellier et~al\mbox{.}}{2019}]%
        {CNIL-Shap-19}
\bibfield{author}{\bibinfo{person}{Régis Chatellier},
  \bibinfo{person}{Geoffrey Delcroix}, \bibinfo{person}{Estelle Hary}, {and}
  \bibinfo{person}{Camille Girard-Chanudet}.} \bibinfo{year}{2019}\natexlab{}.
\newblock \bibinfo{title}{Shaping Choices in the Digital World}.
\newblock
\newblock
\newblock
\shownote{\url{https://linc.cnil.fr/sites/default/files/atoms/files/cnil_ip_report_06_shaping_choices_in_the_digital_world.pdf}.}


\bibitem[\protect\citeauthoryear{Chaturvedi, Poria, and Cambria}{Chaturvedi
  et~al\mbox{.}}{2018}]%
        {ChaturvediPC18}
\bibfield{author}{\bibinfo{person}{Iti Chaturvedi}, \bibinfo{person}{Soujanya
  Poria}, {and} \bibinfo{person}{Erik Cambria}.}
  \bibinfo{year}{2018}\natexlab{}.
\newblock \showarticletitle{Sentiment Analysis, Basic Tasks of}.
\newblock In \bibinfo{booktitle}{\emph{Encyclopedia of Social Network Analysis
  and Mining, 2nd Edition}}, \bibfield{editor}{\bibinfo{person}{Reda Alhajj}
  {and} \bibinfo{person}{Jon~G. Rokne}} (Eds.). \bibinfo{publisher}{Springer}.
\newblock
\urldef\tempurl%
\url{https://doi.org/10.1007/978-1-4939-7131-2\_110159}
\showDOI{\tempurl}


\bibitem[\protect\citeauthoryear{Chivukula, Gray, and Brier}{Chivukula
  et~al\mbox{.}}{2019}]%
        {chivukula-design}
\bibfield{author}{\bibinfo{person}{Shruthi~Sai Chivukula},
  \bibinfo{person}{Colin~M. Gray}, {and} \bibinfo{person}{Jason~A. Brier}.}
  \bibinfo{year}{2019}\natexlab{}.
\newblock \bibinfo{booktitle}{\emph{Analyzing Value Discovery in Design
  Decisions Through Ethicography}}.
\newblock \bibinfo{publisher}{Association for Computing Machinery},
  \bibinfo{address}{New York, NY, USA}, \bibinfo{pages}{1–12}.
\newblock
\showISBNx{9781450359702}
\urldef\tempurl%
\url{https://doi.org/10.1145/3290605.3300307}
\showURL{%
\tempurl}


\bibitem[\protect\citeauthoryear{Chromik, V\"{o}lkel, Eiband, and
  Buschek}{Chromik et~al\mbox{.}}{2019}]%
        {Chromik2019}
\bibfield{author}{\bibinfo{person}{Michael Chromik},
  \bibinfo{person}{Sarah~Theres V\"{o}lkel}, \bibinfo{person}{Malin Eiband},
  {and} \bibinfo{person}{Daniel Buschek}.} \bibinfo{year}{2019}\natexlab{}.
\newblock \showarticletitle{Dark Patterns of Explainability, Transparency, and
  User Control for Intelligent Systems}. In \bibinfo{booktitle}{\emph{Joint
  Proceedings of the ACM IUI 2019 Workshops co-located with the 24th ACM
  Conference on Intelligent User Interfaces (ACM IUI 2019) Los Angeles, USA,
  March 20, 2019}}, \bibfield{editor}{\bibinfo{person}{Christoph Trattner},
  \bibinfo{person}{Denis Parra}, {and} \bibinfo{person}{Nathalie Riche}}
  (Eds.). \bibinfo{publisher}{CEUR},
  \bibinfo{address}{http://ceur-ws.org/Vol-2327/IUI19WS-ExSS2019-7.pdf},
  \bibinfo{pages}{93--104}.
\newblock


\bibitem[\protect\citeauthoryear{Cialdini}{Cialdini}{2001}]%
        {Cialdini-persuasion}
\bibfield{author}{\bibinfo{person}{Robert~B. Cialdini}.}
  \bibinfo{year}{2001}\natexlab{}.
\newblock \showarticletitle{The Science of Persuasion}.
\newblock \bibinfo{journal}{\emph{Scientific American}} \bibinfo{volume}{284},
  \bibinfo{number}{2} (\bibinfo{year}{2001}), \bibinfo{pages}{76--81}.
\newblock
\showISSN{00368733, 19467087}
\urldef\tempurl%
\url{http://www.jstor.org/stable/26059056}
\showURL{%
\tempurl}


\bibitem[\protect\citeauthoryear{Curley, O'Sullivan, Gordon, Tierney, and
  Stavrakakis}{Curley et~al\mbox{.}}{2021}]%
        {CurleyOGTS2021}
\bibfield{author}{\bibinfo{person}{Andrea Curley}, \bibinfo{person}{Dympna
  O'Sullivan}, \bibinfo{person}{Damian Gordon}, \bibinfo{person}{Brendan
  Tierney}, {and} \bibinfo{person}{Ioannis Stavrakakis}.}
  \bibinfo{year}{2021}\natexlab{}.
\newblock \showarticletitle{The Design of a Framework for the Detection of
  Web-Based Dark Patterns}. In \bibinfo{booktitle}{\emph{The Design of a
  Framework for the Detection of Web-Based Dark Patterns}}.
  \bibinfo{publisher}{online}.
\newblock
\newblock
\shownote{\url{https://arrow.tudublin.ie/ascnetcon/3/}.}


\bibitem[\protect\citeauthoryear{Degeling, Utz, Lentzsch, Hosseini, Schaub, and
  Holz}{Degeling et~al\mbox{.}}{2019}]%
        {Degeling2019WeVY}
\bibfield{author}{\bibinfo{person}{Martin Degeling}, \bibinfo{person}{Christine
  Utz}, \bibinfo{person}{Christopher Lentzsch}, \bibinfo{person}{Henry
  Hosseini}, \bibinfo{person}{F. Schaub}, {and} \bibinfo{person}{T. Holz}.}
  \bibinfo{year}{2019}\natexlab{}.
\newblock \showarticletitle{We Value Your Privacy ... Now Take Some Cookies:
  Measuring the GDPR's Impact on Web Privacy}.
\newblock \bibinfo{journal}{\emph{ArXiv}}  \bibinfo{volume}{abs/1808.05096}
  (\bibinfo{year}{2019}).
\newblock


\bibitem[\protect\citeauthoryear{Di~Geronimo, Braz, Fregnan, Palomba, and
  Bacchelli}{Di~Geronimo et~al\mbox{.}}{2020}]%
        {geronimo}
\bibfield{author}{\bibinfo{person}{Linda Di~Geronimo}, \bibinfo{person}{Larissa
  Braz}, \bibinfo{person}{Enrico Fregnan}, \bibinfo{person}{Fabio Palomba},
  {and} \bibinfo{person}{Alberto Bacchelli}.} \bibinfo{year}{2020}\natexlab{}.
\newblock \bibinfo{booktitle}{\emph{UI Dark Patterns and Where to Find Them: A
  Study on Mobile Applications and User Perception}}.
\newblock \bibinfo{publisher}{Association for Computing Machinery},
  \bibinfo{address}{New York, NY, USA}, \bibinfo{pages}{1–14}.
\newblock
\showISBNx{9781450367080}
\urldef\tempurl%
\url{https://doi.org/10.1145/3313831.3376600}
\showURL{%
\tempurl}


\bibitem[\protect\citeauthoryear{Ducato and Marique}{Ducato and
  Marique}{2019}]%
        {DucatoE2019}
\bibfield{author}{\bibinfo{person}{Rossana Ducato} {and}
  \bibinfo{person}{Enguerrand Marique}.} \bibinfo{year}{2019}\natexlab{}.
\newblock \bibinfo{title}{Come to the Dark Side: We Have Patterns. Choice
  Architecture and Design for (Un)Informed Consent}.
\newblock
\newblock
\urldef\tempurl%
\url{https://ssrn.com/abstract=3365952}
\showURL{%
\tempurl}


\bibitem[\protect\citeauthoryear{et~Libertés}{et~Libertés}{2020}]%
        {CNIL-draft-rec-cookies-2020}
\bibfield{author}{\bibinfo{person}{Commission Nationale~Informatique et
  Libertés}.} \bibinfo{year}{2020}\natexlab{}.
\newblock \bibinfo{title}{On the practical procedures for collecting the
  consent provided for in article 82 of the french data protection act,
  concerning operations of storing or gaining access to information in the
  terminal equipment of a user (recommendation ``cookies and other
  trackers'')}.
\newblock
\newblock
\newblock
\shownote{\url{https://www.cnil.fr/sites/default/files/atoms/files/draft_recommendation_cookies_and_other_trackers_en.pdf}.}


\bibitem[\protect\citeauthoryear{Forbrukerr{\aa}det}{Forbrukerr{\aa}det}{2018}]%
        {ForbrukerradeNO}
\bibfield{author}{\bibinfo{person}{Norway Forbrukerr{\aa}det}.}
  \bibinfo{year}{2018}\natexlab{}.
\newblock \bibinfo{booktitle}{\emph{Deceived by design: How tech companies use
  dark patterns to discourage us from exercising our rights to privacy}}.
\newblock Forbrukerr{\aa}det, Norway.
\newblock
\urldef\tempurl%
\url{https://fil.forbrukerradet.no/wp-content/uploads/2018/06/2018-06-27-deceived-by-design-final.pdf}
\showURL{%
\tempurl}


\bibitem[\protect\citeauthoryear{Friedman}{Friedman}{2001}]%
        {friedman_greedy_2001}
\bibfield{author}{\bibinfo{person}{Jerome~H. Friedman}.}
  \bibinfo{year}{2001}\natexlab{}.
\newblock \showarticletitle{Greedy function approximation: {A} gradient
  boosting machine.}
\newblock \bibinfo{journal}{\emph{The Annals of Statistics}}
  \bibinfo{volume}{29}, \bibinfo{number}{5} (\bibinfo{year}{2001}),
  \bibinfo{pages}{1189 -- 1232}.
\newblock
\urldef\tempurl%
\url{https://doi.org/10.1214/aos/1013203451}
\showDOI{\tempurl}
\newblock
\shownote{Publisher: Institute of Mathematical Statistics.}


\bibitem[\protect\citeauthoryear{Fritsch}{Fritsch}{2017}]%
        {Fritsch2017}
\bibfield{author}{\bibinfo{person}{Lothar Fritsch}.}
  \bibinfo{year}{2017}\natexlab{}.
\newblock \showarticletitle{Privacy dark patterns in identity management}. In
  \bibinfo{booktitle}{\emph{Open Identity Summit 2017}},
  \bibfield{editor}{\bibinfo{person}{Lothar Fritsch}, \bibinfo{person}{Heiko
  Roßnagel}, {and} \bibinfo{person}{Detlef Hühnlein}} (Eds.).
  \bibinfo{publisher}{Gesellschaft für Informatik, Bonn},
  \bibinfo{pages}{93--104}.
\newblock


\bibitem[\protect\citeauthoryear{Gra{\ss}l, Schraffenberger,
  Zuiderveen~Borgesius, and Buijzen}{Gra{\ss}l et~al\mbox{.}}{2021}]%
        {Grassl2021}
\bibfield{author}{\bibinfo{person}{Paul Gra{\ss}l}, \bibinfo{person}{Hanna
  Schraffenberger}, \bibinfo{person}{Frederik Zuiderveen~Borgesius}, {and}
  \bibinfo{person}{Moniek Buijzen}.} \bibinfo{year}{2021}\natexlab{}.
\newblock \showarticletitle{Dark and Bright Patterns in Cookie Consent
  Requests}.
\newblock \bibinfo{journal}{\emph{Journal of Digital Social Research}}
  \bibinfo{volume}{3}, \bibinfo{number}{1} (\bibinfo{date}{Feb.}
  \bibinfo{year}{2021}), \bibinfo{pages}{1--38}.
\newblock
\urldef\tempurl%
\url{https://doi.org/10.33621/jdsr.v3i1.54}
\showDOI{\tempurl}


\bibitem[\protect\citeauthoryear{Gray and Chivukula}{Gray and
  Chivukula}{2019}]%
        {Gray-chivukula-mediationUX}
\bibfield{author}{\bibinfo{person}{Colin~M. Gray} {and}
  \bibinfo{person}{Shruthi~Sai Chivukula}.} \bibinfo{year}{2019}\natexlab{}.
\newblock \bibinfo{booktitle}{\emph{Ethical Mediation in UX Practice}}.
\newblock \bibinfo{publisher}{Association for Computing Machinery},
  \bibinfo{address}{New York, NY, USA}, \bibinfo{pages}{1–11}.
\newblock
\showISBNx{9781450359702}
\urldef\tempurl%
\url{https://doi.org/10.1145/3290605.3300408}
\showURL{%
\tempurl}


\bibitem[\protect\citeauthoryear{Gray, Kou, Battles, Hoggatt, and Toombs}{Gray
  et~al\mbox{.}}{2018}]%
        {gray_dark_2018}
\bibfield{author}{\bibinfo{person}{Colin~M. Gray}, \bibinfo{person}{Yubo Kou},
  \bibinfo{person}{Bryan Battles}, \bibinfo{person}{Joseph Hoggatt}, {and}
  \bibinfo{person}{Austin~L. Toombs}.} \bibinfo{year}{2018}\natexlab{}.
\newblock \showarticletitle{The {Dark} ({Patterns}) {Side} of {UX} {Design}}.
  In \bibinfo{booktitle}{\emph{Proceedings of the 2018 {CHI} {Conference} on
  {Human} {Factors} in {Computing} {Systems} - {CHI} '18}}.
  \bibinfo{publisher}{ACM Press}, \bibinfo{address}{Montreal QC, Canada},
  \bibinfo{pages}{1--14}.
\newblock
\showISBNx{978-1-4503-5620-6}
\urldef\tempurl%
\url{https://doi.org/10.1145/3173574.3174108}
\showDOI{\tempurl}


\bibitem[\protect\citeauthoryear{Gray, Santos, Bielova, Toth, and
  Clifford}{Gray et~al\mbox{.}}{2021}]%
        {gray-legalreq}
\bibfield{author}{\bibinfo{person}{Colin~M. Gray}, \bibinfo{person}{Cristiana
  Santos}, \bibinfo{person}{Nataliia Bielova}, \bibinfo{person}{Michael Toth},
  {and} \bibinfo{person}{Damian Clifford}.} \bibinfo{year}{2021}\natexlab{}.
\newblock \bibinfo{booktitle}{\emph{Dark Patterns and the Legal Requirements of
  Consent Banners: An Interaction Criticism Perspective}}.
\newblock \bibinfo{publisher}{Association for Computing Machinery},
  \bibinfo{address}{New York, NY, USA}.
\newblock
\showISBNx{9781450380966}
\urldef\tempurl%
\url{https://doi.org/10.1145/3411764.3445779}
\showURL{%
\tempurl}


\bibitem[\protect\citeauthoryear{Graßl, Schraffenberger, Zuiderveen~Borgesius,
  and Buijzen}{Graßl et~al\mbox{.}}{2021}]%
        {Grassl2021dark}
\bibfield{author}{\bibinfo{person}{Paul Graßl}, \bibinfo{person}{Hanna
  Schraffenberger}, \bibinfo{person}{Frederik Zuiderveen~Borgesius}, {and}
  \bibinfo{person}{Moniek Buijzen}.} \bibinfo{year}{2021}\natexlab{}.
\newblock \showarticletitle{Dark and Bright Patterns in Cookie Consent
  Requests}.
\newblock \bibinfo{journal}{\emph{Journal of Digital Social Research}}
  \bibinfo{volume}{3}, \bibinfo{number}{1} (\bibinfo{date}{Feb}
  \bibinfo{year}{2021}), \bibinfo{pages}{1–38}.
\newblock
\urldef\tempurl%
\url{https://doi.org/10.33621/jdsr.v3i1.54}
\showDOI{\tempurl}


\bibitem[\protect\citeauthoryear{Hausner and Gertz}{Hausner and Gertz}{2021}]%
        {Hausner2021}
\bibfield{author}{\bibinfo{person}{Philip Hausner} {and}
  \bibinfo{person}{Michael Gertz}.} \bibinfo{year}{2021}\natexlab{}.
\newblock \showarticletitle{Dark Patterns in the Interaction with Cookie
  Banners}.
\newblock \bibinfo{journal}{\emph{CoRR}}  \bibinfo{volume}{abs/2103.14956}
  (\bibinfo{year}{2021}).
\newblock
\showeprint[arxiv]{2103.14956}
\urldef\tempurl%
\url{https://arxiv.org/abs/2103.14956}
\showURL{%
\tempurl}


\bibitem[\protect\citeauthoryear{Hils, Woods, and B\"{o}hme}{Hils
  et~al\mbox{.}}{2020}]%
        {hils-2020}
\bibfield{author}{\bibinfo{person}{Maximilian Hils}, \bibinfo{person}{Daniel~W.
  Woods}, {and} \bibinfo{person}{Rainer B\"{o}hme}.}
  \bibinfo{year}{2020}\natexlab{}.
\newblock \showarticletitle{Measuring the Emergence of Consent Management on
  the Web}. In \bibinfo{booktitle}{\emph{Proceedings of the ACM Internet
  Measurement Conference}} (Virtual Event, USA) \emph{(\bibinfo{series}{IMC
  '20})}. \bibinfo{publisher}{Association for Computing Machinery},
  \bibinfo{address}{New York, NY, USA}, \bibinfo{pages}{317–332}.
\newblock
\showISBNx{9781450381383}
\urldef\tempurl%
\url{https://doi.org/10.1145/3419394.3423647}
\showDOI{\tempurl}


\bibitem[\protect\citeauthoryear{Human and Cech}{Human and Cech}{2021}]%
        {Soheil2021}
\bibfield{author}{\bibinfo{person}{Soheil Human} {and} \bibinfo{person}{Florian
  Cech}.} \bibinfo{year}{2021}\natexlab{}.
\newblock \showarticletitle{A Human-Centric Perspective on Digital Consenting:
  The Case of GAFAM}. In \bibinfo{booktitle}{\emph{Human Centred Intelligent
  Systems}}, \bibfield{editor}{\bibinfo{person}{Alfred Zimmermann},
  \bibinfo{person}{Robert~J. Howlett}, {and} \bibinfo{person}{Lakhmi~C. Jain}}
  (Eds.). \bibinfo{publisher}{Springer Singapore},
  \bibinfo{address}{Singapore}, \bibinfo{pages}{139--159}.
\newblock
\showISBNx{978-981-15-5784-2}


\bibitem[\protect\citeauthoryear{Infromation}{Infromation}{2018}]%
        {CCPA}
\bibfield{author}{\bibinfo{person}{California~Legislative Infromation}.}
  \bibinfo{year}{2018}\natexlab{}.
\newblock \bibinfo{booktitle}{\emph{Assembly Bill No. 375, CHAPTER 55,
  Legislative Councel's Digest}}.
\newblock The state of California.
\newblock
\urldef\tempurl%
\url{https://leginfo.legislature.ca.gov/faces/billTextClient.xhtml?bill_id=201720180AB375}
\showURL{%
\tempurl}


\bibitem[\protect\citeauthoryear{Johanna~Gunawan}{Johanna~Gunawan}{2021}]%
        {Gunawan-comparativestudy2021}
\bibfield{author}{\bibinfo{person}{David~Choffnes Johanna~Gunawan,
  Amogh~Pradeep}.} \bibinfo{year}{2021}\natexlab{}.
\newblock \bibinfo{booktitle}{\emph{A Comparative Study of Dark Patterns Across
  Mobile and Web Modalities}}.
\newblock \bibinfo{publisher}{Proc. ACM Hum.-Comput. Interact. 5, CSCW2,
  Article 377}.
\newblock
\urldef\tempurl%
\url{https://doi.org/10.1145/3479521}
\showURL{%
\tempurl}


\bibitem[\protect\citeauthoryear{Kampanos and Shahandashti}{Kampanos and
  Shahandashti}{2021}]%
        {kampanos2021accept}
\bibfield{author}{\bibinfo{person}{Georgios Kampanos} {and}
  \bibinfo{person}{Siamak~F. Shahandashti}.} \bibinfo{year}{2021}\natexlab{}.
\newblock \bibinfo{title}{Accept All: The Landscape of Cookie Banners in Greece
  and the UK}.
\newblock
\newblock
\showeprint[arxiv]{2104.05750}~[cs.CR]


\bibitem[\protect\citeauthoryear{Karegar, Pettersson, and
  Fischer-H\"{u}bner}{Karegar et~al\mbox{.}}{2020}]%
        {Karegar-et-al}
\bibfield{author}{\bibinfo{person}{Farzaneh Karegar},
  \bibinfo{person}{John~S\"{o}ren Pettersson}, {and} \bibinfo{person}{Simone
  Fischer-H\"{u}bner}.} \bibinfo{year}{2020}\natexlab{}.
\newblock \showarticletitle{The Dilemma of User Engagement in Privacy Notices:
  Effects of Interaction Modes and Habituation on User Attention}.
\newblock \bibinfo{journal}{\emph{ACM Trans. Priv. Secur.}}
  \bibinfo{volume}{23}, Article \bibinfo{articleno}{5} (\bibinfo{year}{2020}).
\newblock


\bibitem[\protect\citeauthoryear{Kretschmer, Pennekamp, and Wehrle}{Kretschmer
  et~al\mbox{.}}{2021}]%
        {Kretschmer-cookiebanners21}
\bibfield{author}{\bibinfo{person}{Michael Kretschmer}, \bibinfo{person}{Jan
  Pennekamp}, {and} \bibinfo{person}{Klaus Wehrle}.}
  \bibinfo{year}{2021}\natexlab{}.
\newblock \showarticletitle{Cookie Banners and Privacy Policies: Measuring the
  Impact of the GDPR on the Web}.
\newblock \bibinfo{journal}{\emph{ACM Trans. Web}} \bibinfo{volume}{15},
  \bibinfo{number}{4}, Article \bibinfo{articleno}{20} (\bibinfo{date}{July}
  \bibinfo{year}{2021}), \bibinfo{numpages}{42}~pages.
\newblock
\showISSN{1559-1131}
\urldef\tempurl%
\url{https://doi.org/10.1145/3466722}
\showDOI{\tempurl}


\bibitem[\protect\citeauthoryear{Kulyk, Hilt, Gerber, and Volkamer}{Kulyk
  et~al\mbox{.}}{2018}]%
        {KulykOksana-cookieperception}
\bibfield{author}{\bibinfo{person}{Oksana Kulyk}, \bibinfo{person}{Annika
  Hilt}, \bibinfo{person}{Nina Gerber}, {and} \bibinfo{person}{Melanie
  Volkamer}.} \bibinfo{year}{2018}\natexlab{}.
\newblock \showarticletitle{{``This website uses cookies"}: Users’
  perceptions and reactions to the cookie disclaimer}.
\newblock In \bibinfo{booktitle}{\emph{European Workshop on Usable Security
  (EuroUSEC)}}.
\newblock
\urldef\tempurl%
\url{https://doi.org/10.14722/eurousec.2018.23012}
\showDOI{\tempurl}


\bibitem[\protect\citeauthoryear{Lloyd}{Lloyd}{1982}]%
        {Lloyd82leastsquares}
\bibfield{author}{\bibinfo{person}{Stuart~P. Lloyd}.}
  \bibinfo{year}{1982}\natexlab{}.
\newblock \showarticletitle{Least squares quantization in pcm}.
\newblock \bibinfo{journal}{\emph{IEEE Transactions on Information Theory}}
  \bibinfo{volume}{28} (\bibinfo{year}{1982}), \bibinfo{pages}{129--137}.
\newblock


\bibitem[\protect\citeauthoryear{Luguri and Strahilevitz}{Luguri and
  Strahilevitz}{2019}]%
        {Luguri2019ShiningAL}
\bibfield{author}{\bibinfo{person}{Jamie~B. Luguri} {and} \bibinfo{person}{L.
  Strahilevitz}.} \bibinfo{year}{2019}\natexlab{}.
\newblock \showarticletitle{Shining a Light on Dark Patterns}.
\newblock \bibinfo{journal}{\emph{Behavioral \& Experimental Economics
  eJournal}} (\bibinfo{year}{2019}).
\newblock


\bibitem[\protect\citeauthoryear{Maier and Harr}{Maier and Harr}{2020}]%
        {Maier2020DarkDP}
\bibfield{author}{\bibinfo{person}{Maximilian Maier} {and}
  \bibinfo{person}{Rikard Harr}.} \bibinfo{year}{2020}\natexlab{}.
\newblock \showarticletitle{Dark Design Patterns : An End-user Perspective}.
\newblock \bibinfo{journal}{\emph{Human Technology}}  \bibinfo{volume}{16}
  (\bibinfo{year}{2020}), \bibinfo{pages}{170--199}.
\newblock


\bibitem[\protect\citeauthoryear{Mathur, Acar, Friedman, Lucherini, Mayer,
  Chetty, and Narayanan}{Mathur et~al\mbox{.}}{2019}]%
        {mathur_dark_2019}
\bibfield{author}{\bibinfo{person}{Arunesh Mathur}, \bibinfo{person}{Gunes
  Acar}, \bibinfo{person}{Michael~J. Friedman}, \bibinfo{person}{Elena
  Lucherini}, \bibinfo{person}{Jonathan Mayer}, \bibinfo{person}{Marshini
  Chetty}, {and} \bibinfo{person}{Arvind Narayanan}.}
  \bibinfo{year}{2019}\natexlab{}.
\newblock \showarticletitle{Dark Patterns at Scale: Findings from a Crawl of
  11K Shopping Websites}.
\newblock \bibinfo{journal}{\emph{Proc. ACM Hum.-Comput. Interact.}}
  \bibinfo{volume}{3}, \bibinfo{number}{CSCW}, Article \bibinfo{articleno}{81}
  (\bibinfo{date}{Nov.} \bibinfo{year}{2019}), \bibinfo{numpages}{32}~pages.
\newblock
\urldef\tempurl%
\url{https://doi.org/10.1145/3359183}
\showDOI{\tempurl}


\bibitem[\protect\citeauthoryear{Mathur, Mayer, and Kshirsagar}{Mathur
  et~al\mbox{.}}{2021}]%
        {Mathur-2021}
\bibfield{author}{\bibinfo{person}{Arunesh Mathur}, \bibinfo{person}{Jonathan
  Mayer}, {and} \bibinfo{person}{Mihir Kshirsagar}.}
  \bibinfo{year}{2021}\natexlab{}.
\newblock \showarticletitle{What Makes a Dark Pattern... Dark? Design
  Attributes, Normative Considerations, and Measurement Methods}. In
  \bibinfo{booktitle}{\emph{ACM Conference on Human Factors in Computing
  Systems}}.
\newblock
\urldef\tempurl%
\url{https://arxiv.org/abs/2101.04843}
\showURL{%
\tempurl}


\bibitem[\protect\citeauthoryear{Matte, Bielova, and Santos}{Matte
  et~al\mbox{.}}{2019}]%
        {MatteBS2019}
\bibfield{author}{\bibinfo{person}{C{\'{e}}lestin Matte},
  \bibinfo{person}{Nataliia Bielova}, {and} \bibinfo{person}{Cristiana
  Santos}.} \bibinfo{year}{2019}\natexlab{}.
\newblock \showarticletitle{Do Cookie Banners Respect my Choice? Measuring
  Legal Compliance of Banners from {IAB} Europe's Transparency and Consent
  Framework}.
\newblock \bibinfo{journal}{\emph{CoRR}}  \bibinfo{volume}{abs/1911.09964}
  (\bibinfo{year}{2019}).
\newblock
\showeprint[arxiv]{1911.09964}
\urldef\tempurl%
\url{http://arxiv.org/abs/1911.09964}
\showURL{%
\tempurl}


\bibitem[\protect\citeauthoryear{Matte, Bielova, and Santos}{Matte
  et~al\mbox{.}}{2020}]%
        {matte_cookie_2020}
\bibfield{author}{\bibinfo{person}{Célestin Matte}, \bibinfo{person}{Nataliia
  Bielova}, {and} \bibinfo{person}{Cristiana Santos}.}
  \bibinfo{year}{2020}\natexlab{}.
\newblock \showarticletitle{Do {Cookie} {Banners} {Respect} my {Choice}?
  {Measuring} {Legal} {Compliance} of {Banners} from {IAB} {Europe}'s
  {Transparency} and {Consent} {Framework}}.
\newblock \bibinfo{journal}{\emph{arXiv:1911.09964 [cs]}} (\bibinfo{date}{Feb.}
  \bibinfo{year}{2020}).
\newblock
\urldef\tempurl%
\url{http://arxiv.org/abs/1911.09964}
\showURL{%
\tempurl}
\newblock
\shownote{arXiv: 1911.09964.}


\bibitem[\protect\citeauthoryear{Nouwens, Liccardi, Veale, Karger, and
  Kagal}{Nouwens et~al\mbox{.}}{2020a}]%
        {NouwensLVKK20}
\bibfield{author}{\bibinfo{person}{Midas Nouwens}, \bibinfo{person}{Ilaria
  Liccardi}, \bibinfo{person}{Michael Veale}, \bibinfo{person}{David Karger},
  {and} \bibinfo{person}{Lalana Kagal}.} \bibinfo{year}{2020}\natexlab{a}.
\newblock \showarticletitle{Dark Patterns after the {GDPR:} Scraping Consent
  Pop-ups and Demonstrating their Influence}.
\newblock \bibinfo{journal}{\emph{CoRR}}  \bibinfo{volume}{abs/2001.02479}
  (\bibinfo{year}{2020}).
\newblock
\showeprint[arxiv]{2001.02479}
\urldef\tempurl%
\url{http://arxiv.org/abs/2001.02479}
\showURL{%
\tempurl}


\bibitem[\protect\citeauthoryear{Nouwens, Liccardi, Veale, Karger, and
  Kagal}{Nouwens et~al\mbox{.}}{2020b}]%
        {nouwens_dark_2020}
\bibfield{author}{\bibinfo{person}{Midas Nouwens}, \bibinfo{person}{Ilaria
  Liccardi}, \bibinfo{person}{Michael Veale}, \bibinfo{person}{David Karger},
  {and} \bibinfo{person}{Lalana Kagal}.} \bibinfo{year}{2020}\natexlab{b}.
\newblock \showarticletitle{Dark {Patterns} after the {GDPR}: {Scraping}
  {Consent} {Pop}-ups and {Demonstrating} their {Influence}}.
\newblock \bibinfo{journal}{\emph{arXiv:2001.02479 [cs]}} (\bibinfo{date}{Jan.}
  \bibinfo{year}{2020}).
\newblock
\urldef\tempurl%
\url{https://doi.org/10.1145/3313831.3376321}
\showDOI{\tempurl}
\newblock
\shownote{arXiv: 2001.02479.}


\bibitem[\protect\citeauthoryear{of~Justice of~the European~Union}{of~Justice
  of~the European~Union}{2019}]%
        {CJEU-Planet49-19}
\bibfield{author}{\bibinfo{person}{Court of~Justice of~the European~Union}.}
  \bibinfo{year}{2019}\natexlab{}.
\newblock \bibinfo{title}{Judgment in Case C-673/17 Bundesverband der
  Verbraucherzentralen und Verbraucherverbände -- Verbraucherzentrale
  Bundesverband eV v Planet49 GmbH}.
\newblock
\newblock
\newblock
\shownote{\url{http://curia.europa.eu/juris/documents.jsf?num=C-673/17}.}


\bibitem[\protect\citeauthoryear{Office}{Office}{2019}]%
        {ICO-Guid-19}
\bibfield{author}{\bibinfo{person}{Information~Commissioner's Office}.}
  \bibinfo{year}{2019}\natexlab{}.
\newblock \bibinfo{title}{Guidance on the use of cookies and similar
  technologies}.
\newblock
\newblock
\newblock
\shownote{\url{https://ico.org.uk/media/for-organisations/guide-to-pecr/guidance-on-the-use-of-cookies-and-similar-technologies-1-0.pdf}.}


\bibitem[\protect\citeauthoryear{on~Digital~Platforms}{on~Digital~Platforms}{2019}]%
        {Stigler-report-2019}
\bibfield{author}{\bibinfo{person}{Stigler~Committee on Digital~Platforms}.}
  \bibinfo{year}{2019}\natexlab{}.
\newblock \bibinfo{title}{Privacy and Data Protection Subcommittee Report}.
\newblock
\newblock
\newblock
\shownote{\url{https://research.chicagobooth.edu/-/media/research/stigler/pdfs/data---report.pdf?la=en&hash=54ABA86A7A50C926458B5D44FBAAB83D673DB412}.}


\bibitem[\protect\citeauthoryear{Parlament and Council}{Parlament and
  Council}{2016}]%
        {GDPR}
\bibfield{author}{\bibinfo{person}{European Parlament} {and}
  \bibinfo{person}{Council}.} \bibinfo{year}{2016}\natexlab{}.
\newblock \bibinfo{booktitle}{\emph{Regulation (EU) 2016/679 of the European
  Parliament and of the Council of 27 April 2016 on the protection of natural
  persons with regard to the processing of personal data and on the free
  movement of such data, and repealing Directive 95/46/EC (General Data
  Protection Regulation)}}.
\newblock EU.
\newblock
\urldef\tempurl%
\url{http://data.europa.eu/eli/reg/2016/679/oj}
\showURL{%
\tempurl}


\bibitem[\protect\citeauthoryear{Pearson}{Pearson}{1901}]%
        {PCA}
\bibfield{author}{\bibinfo{person}{Karl Pearson}.}
  \bibinfo{year}{1901}\natexlab{}.
\newblock \showarticletitle{On lines and planes of closest fit to systems of
  points in space}.
\newblock \bibinfo{journal}{\emph{The London, Edinburgh, and Dublin
  Philosophical Magazine and Journal of Science}} \bibinfo{volume}{2},
  \bibinfo{number}{11} (\bibinfo{year}{1901}), \bibinfo{pages}{559--572}.
\newblock
\urldef\tempurl%
\url{https://doi.org/10.1080/14786440109462720}
\showDOI{\tempurl}
\showeprint{https://doi.org/10.1080/14786440109462720}


\bibitem[\protect\citeauthoryear{Sanchez-Rola, Dell'Amico, Kotzias, Balzarotti,
  Bilge, Vervier, and Santos}{Sanchez-Rola et~al\mbox{.}}{2019}]%
        {Sanchez-Rola-cookies2019}
\bibfield{author}{\bibinfo{person}{Iskander Sanchez-Rola},
  \bibinfo{person}{Matteo Dell'Amico}, \bibinfo{person}{Platon Kotzias},
  \bibinfo{person}{Davide Balzarotti}, \bibinfo{person}{Leyla Bilge},
  \bibinfo{person}{Pierre-Antoine Vervier}, {and} \bibinfo{person}{Igor
  Santos}.} \bibinfo{year}{2019}\natexlab{}.
\newblock \showarticletitle{Can I Opt Out Yet? GDPR and the Global Illusion of
  Cookie Control}. In \bibinfo{booktitle}{\emph{Proceedings of the 2019 ACM
  Asia Conference on Computer and Communications Security}} (Auckland, New
  Zealand) \emph{(\bibinfo{series}{Asia CCS '19})}.
  \bibinfo{publisher}{Association for Computing Machinery},
  \bibinfo{address}{New York, NY, USA}, \bibinfo{pages}{340–351}.
\newblock
\showISBNx{9781450367523}
\urldef\tempurl%
\url{https://doi.org/10.1145/3321705.3329806}
\showDOI{\tempurl}


\bibitem[\protect\citeauthoryear{Santos, Bielova, and Matte}{Santos
  et~al\mbox{.}}{2020}]%
        {Sant-etal-20-TechReg}
\bibfield{author}{\bibinfo{person}{Cristiana Santos}, \bibinfo{person}{Nataliia
  Bielova}, {and} \bibinfo{person}{C{é}lestin Matte}.}
  \bibinfo{year}{2020}\natexlab{}.
\newblock \showarticletitle{Are cookie banners indeed compliant with the law?
  Deciphering {EU} legal requirements on consent and technical means to verify
  compliance of cookie banners}.
\newblock \bibinfo{journal}{\emph{Technology and Regulation}}
  (\bibinfo{year}{2020}), \bibinfo{pages}{91--135}.
\newblock
\urldef\tempurl%
\url{https://doi.org/10.26116/techreg.2020.009}
\showURL{%
\tempurl}


\bibitem[\protect\citeauthoryear{Santos, Gangemi, and Alam}{Santos
  et~al\mbox{.}}{2017}]%
        {Santos2017DetectingAE}
\bibfield{author}{\bibinfo{person}{Cristiana Santos}, \bibinfo{person}{Aldo
  Gangemi}, {and} \bibinfo{person}{Mehwish Alam}.}
  \bibinfo{year}{2017}\natexlab{}.
\newblock \showarticletitle{Detecting and Editing Privacy Policy Pitfalls on
  the Web}. In \bibinfo{booktitle}{\emph{TERECOM@JURIX}}.
\newblock


\bibitem[\protect\citeauthoryear{Santos, Nouwens, Toth, Bielova, and
  Roca}{Santos et~al\mbox{.}}{2021}]%
        {Santos-cmp21}
\bibfield{author}{\bibinfo{person}{Cristiana Santos}, \bibinfo{person}{Midas
  Nouwens}, \bibinfo{person}{Michael Toth}, \bibinfo{person}{Nataliia Bielova},
  {and} \bibinfo{person}{Vincent Roca}.} \bibinfo{year}{2021}\natexlab{}.
\newblock \showarticletitle{Consent Management Platforms under the GDPR:
  processors and/or controllers?}. In \bibinfo{booktitle}{\emph{Gruschka N.,
  Antunes L.F.C., Rannenberg K., Drogkaris P. (eds) Privacy Technologies and
  Policy. APF 2021. Lecture Notes in Computer Science, vol 12703. Springer}}.
\newblock


\bibitem[\protect\citeauthoryear{Slavkovik, Stachl, Pitman, and
  Askonas}{Slavkovik et~al\mbox{.}}{2021}]%
        {aies-2021}
\bibfield{author}{\bibinfo{person}{Marija Slavkovik}, \bibinfo{person}{Clemens
  Stachl}, \bibinfo{person}{Caroline Pitman}, {and} \bibinfo{person}{Jonathan
  Askonas}.} \bibinfo{year}{2021}\natexlab{}.
\newblock \showarticletitle{Digital Voodoo Dolls}. In
  \bibinfo{booktitle}{\emph{Proceedings of the 2021 AAAI/ACM Conference on AI,
  Ethics, and Society, May 19--21, 2021, Virtual Event, USA}}.
\newblock
\urldef\tempurl%
\url{https://arxiv.org/abs/2105.02738}
\showURL{%
\tempurl}
\newblock
\shownote{preprint available at ArXiv.org.}


\bibitem[\protect\citeauthoryear{Soe, Nordberg, Guribye, and Slavkovik}{Soe
  et~al\mbox{.}}{2020}]%
        {SoeNGS20}
\bibfield{author}{\bibinfo{person}{Than~Htut Soe}, \bibinfo{person}{Oda~Elise
  Nordberg}, \bibinfo{person}{Frode Guribye}, {and} \bibinfo{person}{Marija
  Slavkovik}.} \bibinfo{year}{2020}\natexlab{}.
\newblock \showarticletitle{Circumvention by design - dark patterns in cookie
  consent for online news outlets}. In \bibinfo{booktitle}{\emph{NordiCHI '20:
  Shaping Experiences, Shaping Society, Proceedings of the 11th Nordic
  Conference on Human-Computer Interaction, Tallinn, Estonia, 25-29 October,
  2020}}, \bibfield{editor}{\bibinfo{person}{David Lamas},
  \bibinfo{person}{Hegle Sarapuu}, \bibinfo{person}{Marta
  L{\'{a}}rusd{\'{o}}ttir}, \bibinfo{person}{Jan Stage}, {and}
  \bibinfo{person}{Carmelo Ardito}} (Eds.). \bibinfo{publisher}{{ACM}},
  \bibinfo{pages}{19:1--19:12}.
\newblock
\urldef\tempurl%
\url{https://doi.org/10.1145/3419249.3420132}
\showDOI{\tempurl}


\bibitem[\protect\citeauthoryear{Strycharz, Smit, Helberger, and {van
  Noort}}{Strycharz et~al\mbox{.}}{2021}]%
        {STRYCHARZ2021106750}
\bibfield{author}{\bibinfo{person}{Joanna Strycharz}, \bibinfo{person}{Edith
  Smit}, \bibinfo{person}{Natali Helberger}, {and} \bibinfo{person}{Guda {van
  Noort}}.} \bibinfo{year}{2021}\natexlab{}.
\newblock \showarticletitle{No to cookies: Empowering impact of technical and
  legal knowledge on rejecting tracking cookies}.
\newblock \bibinfo{journal}{\emph{Computers in Human Behavior}}
  \bibinfo{volume}{120} (\bibinfo{year}{2021}), \bibinfo{pages}{106750}.
\newblock
\showISSN{0747-5632}
\urldef\tempurl%
\url{https://doi.org/10.1016/j.chb.2021.106750}
\showDOI{\tempurl}


\bibitem[\protect\citeauthoryear{Szpunar}{Szpunar}{2019}]%
        {Adv-gen-Szpunar-2019}
\bibfield{author}{\bibinfo{person}{Advocate~General Szpunar}.}
  \bibinfo{year}{2019}\natexlab{}.
\newblock \bibinfo{title}{{Opinion of Advocate General Szpunar in Case
  C-673/17, ECLI:EU:C:2019:246 -- Planet49 GmbH v Bundes verbandder
  Verbraucherzentralen und Verbraucherverbände–Verbraucherzentrale
  Bundesverbande}}.
\newblock
\newblock
\newblock
\shownote{\url{https://eur-lex.europa.eu/legal-content/EN/TXT/?uri=CELEX:62017CC0673}.}


\bibitem[\protect\citeauthoryear{the Council of~the European~Union}{the Council
  of~the European~Union}{2009}]%
        {ePD-09}
\bibfield{author}{\bibinfo{person}{The European Parliament~\& the Council
  of~the European~Union}.} \bibinfo{year}{2009}\natexlab{}.
\newblock \bibinfo{title}{Directive 2009/136/EC of the European Parliament and
  of the Council}.
\newblock
\newblock
\newblock
\shownote{\url{https://eur-lex.europa.eu/LexUriServ/LexUriServ.do?uri=OJ:L:2009:337:0011:0036:En:PDF}.}


\bibitem[\protect\citeauthoryear{Utz, Degeling, Fahl, Schaub, and Holz}{Utz
  et~al\mbox{.}}{2019}]%
        {utz_informed_2019}
\bibfield{author}{\bibinfo{person}{Christine Utz}, \bibinfo{person}{Martin
  Degeling}, \bibinfo{person}{Sascha Fahl}, \bibinfo{person}{Florian Schaub},
  {and} \bibinfo{person}{Thorsten Holz}.} \bibinfo{year}{2019}\natexlab{}.
\newblock \showarticletitle{({Un})informed {Consent}: {Studying} {GDPR}
  {Consent} {Notices} in the {Field}}.
\newblock \bibinfo{journal}{\emph{Proceedings of the 2019 ACM SIGSAC Conference
  on Computer and Communications Security}} (\bibinfo{date}{Nov.}
  \bibinfo{year}{2019}), \bibinfo{pages}{973--990}.
\newblock
\urldef\tempurl%
\url{https://doi.org/10.1145/3319535.3354212}
\showDOI{\tempurl}
\newblock
\shownote{arXiv: 1909.02638.}


\bibitem[\protect\citeauthoryear{Waldman}{Waldman}{2020}]%
        {WALDMAN2020105}
\bibfield{author}{\bibinfo{person}{Ari~Ezra Waldman}.}
  \bibinfo{year}{2020}\natexlab{}.
\newblock \showarticletitle{Cognitive biases, dark patterns, and the ‘privacy
  paradox’}.
\newblock \bibinfo{journal}{\emph{Current Opinion in Psychology}}
  \bibinfo{volume}{31} (\bibinfo{year}{2020}), \bibinfo{pages}{105--109}.
\newblock
\showISSN{2352-250X}
\urldef\tempurl%
\url{https://doi.org/10.1016/j.copsyc.2019.08.025}
\showDOI{\tempurl}
\newblock
\shownote{Privacy and Disclosure, Online and in Social Interactions.}


\bibitem[\protect\citeauthoryear{Zagal, Bj{\"o}rk, and Lewis}{Zagal
  et~al\mbox{.}}{2013}]%
        {Zagal2013DarkPI}
\bibfield{author}{\bibinfo{person}{J. Zagal}, \bibinfo{person}{Staffan
  Bj{\"o}rk}, {and} \bibinfo{person}{Chris Lewis}.}
  \bibinfo{year}{2013}\natexlab{}.
\newblock \showarticletitle{Dark patterns in the design of games}. In
  \bibinfo{booktitle}{\emph{FDG}}.
\newblock


\end{thebibliography}
\clearpage
\appendix
\section{Background: dark patterns in cookie banners and their impact in user decision making}\label{sec:background}
In this section we give the preliminary definitions of concepts that we use throughout the paper.

\noindent
\textbf{Dark patterns.}
The term dark patterns \citep{gray_dark_2018} has been coined\footnote{The neologism, dark pattern, was coined by user experience designer Harry Brignull in 2010.} to identify ``instances where designers use their knowledge of human behavior (e.g., psychology) and the desires of end users to implement deceptive functionality that is not in the user's best interest'' \citep{gray_dark_2018}. Throughout this work, we use the concept ``dark pattern''
to refer to types of  UI dark designs that have been documented and applied to cookie banners on websites~\citep{gray_dark_2018, nouwens_dark_2020}. Gray et. al. define five park patterns categories from:  nagging, obstruction, sneaking, interface interference, and forced action.  The description of each of these patterns as given  in Table~\ref{tab:dark}.

 \begin{table}[h!] 
    \small
    \centering
    \resizebox{\textwidth}{!}{
    \begin{tabular}{a|p{10cm}}
  \rowcolor{Gray}  Name  & Description\\\hline
   Nagging     &  A minor redirection of expected functionality that may persist over one or more interactions. Nagging often manifests as a repeated intrusion during normal interaction, where the user's desired task is interrupted one or more times by other tasks not directly related to the one the user is focusing on. \\\hline
     Obstruction    & Impending a task flow, making an interaction more difficult than it inherently needs to be with the intent to dissuade an action. Obstruction often manifests as a major barrier to a particular task the user may want to accomplish.\\\hline
     Sneaking & An attempt to hide, disguise, or delay the divulging of information that has relevance to the user. Sneaking often occurs in order to make the user perform an action they may object to if they had the knowledge.\\\hline
     Interface $\;\;$ interference & Any manipulation of the user interface that privileges specific actions over others, thereby confusing the user or limiting discoverability of important action possibilities. Interface interference manifests as numerous individual visual and interactive deceptions.\\\hline
     Forced action & Any situation in which users are required to perform a specific action to access (or continue to access) specific functionality. This action may manifest as a required step to complete a process, or may appear disguised as an option that the user will greatly benefit from.
    \end{tabular}}
    \caption{Dark pattern types of \citep{gray_dark_2018} and their definitions}
    \label{tab:dark}
\end{table}

\vspace{1mm} 

\noindent
\textbf{Legal requirements for consent in cookie banners.}
\textit{Consent} is defined in Article 4(11) and complemented by Articles 6 and 7 of the GDPR which state  that for consent to be valid, it must satisfy the following seven requirements: it must be prior, freely given, specific, informed, unambiguous,
must be readable and accessible, and finally,  revocable. 
%
%
\textit{Unambiguous} means that consent must be given through an active behavior of the user through which she indicates acceptance or refusal to online tracking. 
Such active behaviors can consist of: \textit{"clicking on a link, or a button, box, image or other content on the entry webpage, or by any other active behavior from which a website operator can unambiguously conclude it means consent"}~\citep{WP208-guidance-cookies}.
Accordingly, silence, pre-ticked boxes or inactivity should not therefore constitute consent and violate such  unambiguous requirement (Recital 32 GDPR, \citep{CJEU-Planet49-19}).
The unambiguous requirement
entails that privacy options (accept, reject, revoke, configure, know more) should be \textit{balanced} (or equitable) (Article 7((4) a contrario, GDPR~\citep{Sant-etal-20-TechReg,ICO-Guid-19}). 
Websites
failing to comply with these GDPR requirements  face fines up to 4\% of their annual revenue
or 20 million euros (Article 83(5,6)).

\medskip

\noindent
\textbf{Impact of dark patters in the user decision making process.}
Deployment of dark patterns by online services are  observed to be effective at
bending people towards choices that are not in their own interest, impacting their decision-making process. 
%
%

\textit{Studies.} The effect of dark patterns is evidenced in a growing studies and experiments.
Nouwens et al.~\citep{nouwens_dark_2020}  ran a user study on 40 participants to assess  the effect that cookie banner design has on  consent and found that there was an approximate 22\% of increase in
acceptance when the opt-out option was “hidden” behind the initial
cookie banner (at least two clicks are needed to opt out).

 Di Geronimo et al.~\citep{geronimo}  coined the term
\textit{ "dark pattern-blindness"} motivating why most respondents in their study were not able to
recognise dark patterns in mobile applications, though when these  were informed of the potential presence of
dark patterns in the context at hand, they became more capable of spotting them.

In the same line, Bhoot et al.~\citep{Bhoot2020TowardsTI}  observed that if the interface is appealing, respondents tend to experience less frustration and hardly notice manipulative attempts. The experiment  shows that
certain design elements can influence people’s capacity of identifying and resisting dark patterns.
 
The study by Utz et al.~\citep{utz_informed_2019} showed that UI design tricks have a very pronounced affect on the decision made by people on whether they will interact with the cookie banners and whether they will accept or reject cookie banners.
 
Maier and Harr~\citep{Maier2020DarkDP} reveal in the respondents answers awareness, annoyance and
resignation, as their participants believed it impossible to avoid
online manipulation, and  acknowledged that the trade-off (free service) outweighs negative consequences.
 
Degeling et al.~\citep{Degeling2019WeVY} studied
how banners design affect users’ choice, and notably finds that
the absence of a “refuse” button on the first layer of the banner
increases positive consent by about 22\%.
 
Luguri and Strahilevitz~\citep{Luguri2019ShiningAL} found
that  dark UI caused participants in their survey and experiment to accept costly service almost four times as often as the same interface without  dark patterns. Their work showed that subtle dark patterns are  easily unnoticed than ostensive ones, and that less-educated people are prone to be influenced than more educated subjects.
 
Kulyk et al.~\citep{KulykOksana-cookieperception}  made an explorative  survey with 150
participants in order to study the perception of such cookie banners among the users, as well as the users’ reactions to
such a disclaimer and factors that influence these reactions. The study  showed that  users tend to have a negative perception of cookie banners, either perceiving it as a nuisance
or as a threat to their privacy,  were distrustful towards textual statements.
 
Finally, 
Bongard-Blanchy et al.~~\citep{Bongard2021manipulated} reveal in their user study  that
users are able to recognize dark patterns, but fail to understand how   manipulative design can concretely harm them. It furthermore hints that a
higher ability to discern manipulative designs is positively related
the capacity to self-protect, though it also finds that the most deceptive dark patterns were harder to be identified by users. The authors convey that wrong mental models (about their risks) and grown habituation to such designs make certain dark patterns harder to spot.

\begin{table}[h!]
 \resizebox{\textwidth}{!}{
\begin{tabular}{|p{7cm}|p{7cm}|} 
 \hline
 Minimum requirements for  GDPR compliance~\citep{nouwens_dark_2020} & GDPR violations~\citep{matte_cookie_2020}   \\
\hhline{==}
\textbf{Consent must be 
unambiguous} - Consent must be a clear affirmative action, such as clicking a button or ticking a box (Article 4(11), Recital 32 GDPR) & \textbf{No way to opt out} - The option to refuse consent is not available  \\ 
 \hline
 \textbf{Accepting all is as easy as rejecting all} - Consent must be easy to give as to withdrawal/refuse (Article 7(3) GDPR) & \textbf{Non-respect of choice} - Consent choice made is not respected in the cookie settings \\ 
  \hline

 \textbf{No pre-ticked boxes} - Consent to 
 tracking purposes must be unchecked by default~\citep{CJEU-Planet49-19} & \textbf{Pre-selected choices} - All vendors or purposes choices should not be preselected \\
  \hline

\textbf{Prior consent} - Consent must be given prior to any data processing (Article 6 (1)(a) GDPR) & \textbf{Consent stored before choice} \\
 \hline
\end{tabular}}
\caption{Comparisons of GDPR violations measures used in ~\citep{matte_cookie_2020, nouwens_dark_2020}}
\label{table:gdpr_conditions_studies}
\end{table}

\textit{Solutions.} In terms of solutions, Graßl et al.~\citep{Grassl2021dark} introduced the term \textit{bright patterns} meaning to redirect  users’ consent decisions towards  privacy-friendly
choices (e.g., pre-selection of "Do not agree" option). 
 However, they also show that even after removing a nudging and manipulative design choice, a form of routinised conditioning could still \emph{persist}, ultimately leading users to behave in a certain way, due to an irreflective default behavior.

\vspace{1mm} 

\textit{Summary.} Building on this body of knowledge made us realize the following: 
i)  user's decision making process is impacted by dark patterns; 
ii)  users  might be aware and recognize them, though they are unable to resist them, and are bound to the immediate trade-off;
iii) some dark patterns are not so easy to be detected by users due to habituation to deceptive design and due to incorrect mental models.
 {As such, automated detection of dark patterns and due reporting to decision-makers (data protection authorities) seems to be the appropriate and neutral intervention  to ease autonomous decision-making and  counteract manipulative designs online}. 

\section{Figures}
 \label{APPENDIX}

\begin{figure}[h]
    \includegraphics[width=0.8\textwidth]{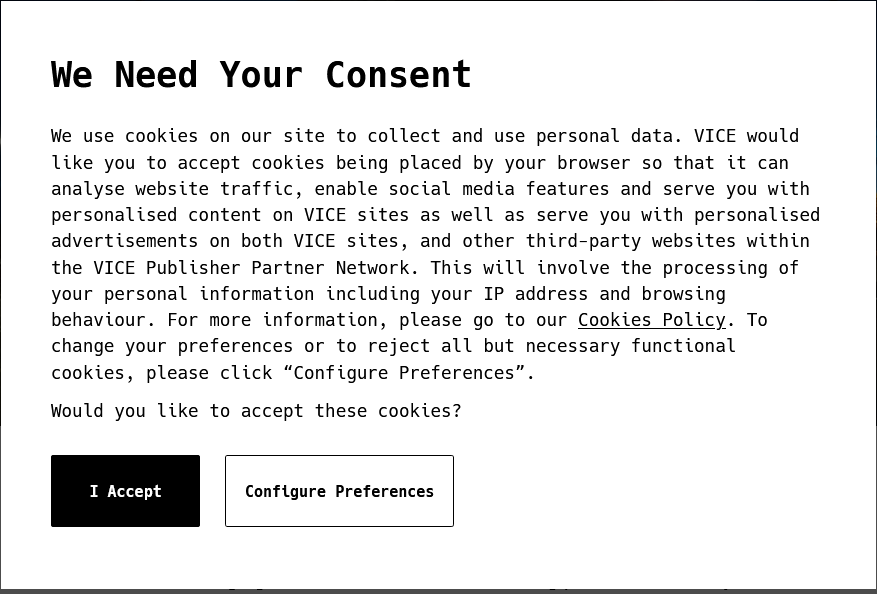}
    \caption{An example of a cookie banner from vice.com retrieved on July 2019} 
    \label{fig:cookieconsent}
\end{figure} 

\begin{figure}[ht]
\begin{subfigure}{.4\textwidth}
  \includegraphics[width=\linewidth]{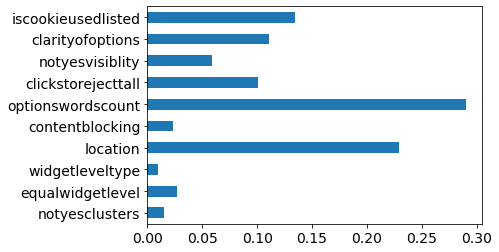}  
  \caption{Nagging}
  \label{fig:sub-first}
\end{subfigure}
\begin{subfigure}{.4\textwidth}
  \includegraphics[width=\linewidth]{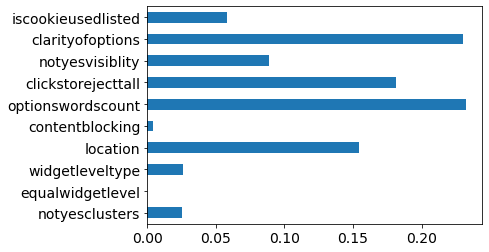}  
  \caption{Obstruction}
  \label{fig:sub-second}
\end{subfigure}
\begin{subfigure}{.4\textwidth}
  \includegraphics[width=\linewidth]{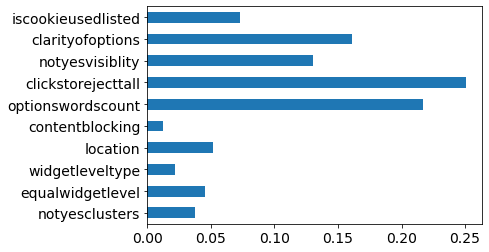}  
  \caption{Sneaking}
  \label{fig:sub-third}
\end{subfigure}
\begin{subfigure}{.4\textwidth}
  \includegraphics[width=\linewidth]{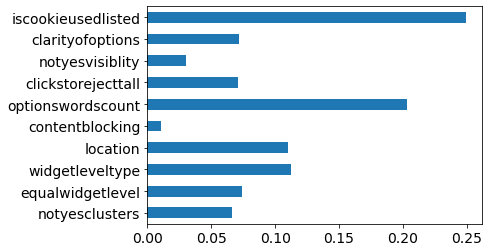}  
  \caption{Interface Interference}
  \label{fig:sub-fourth}
\end{subfigure}

\begin{subfigure}{.4\textwidth}
  \includegraphics[width=\linewidth]{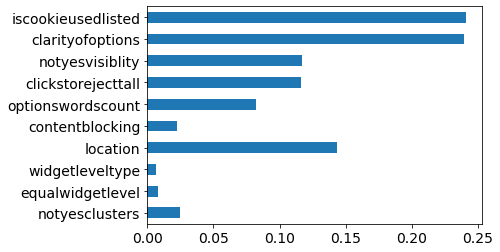}  
  \caption{Forced Action}
  \label{fig:sub-figth}
  \end{subfigure}
\caption{Feature importance comparisons of different dark patterns}
\label{fig:featureimportance}
\end{figure}

\begin{figure}[h!]
\begin{subfigure}{.45\textwidth}
  \centering
  \includegraphics[width=1\linewidth]{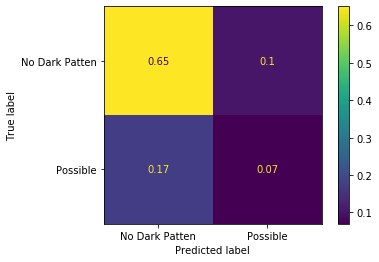}  
  \caption{Nagging}
  \label{fig:confusion-nagging}
\end{subfigure}
\begin{subfigure}{.45\textwidth}
  \centering
  \includegraphics[width=1\linewidth]{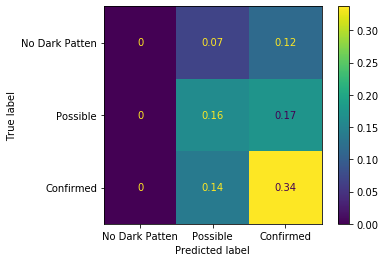}  
  \caption{Obstruction}
  \label{fig:confusion-obstruction}
\end{subfigure}
\begin{subfigure}{.45\textwidth}
  \centering
  \includegraphics[width=1\linewidth]{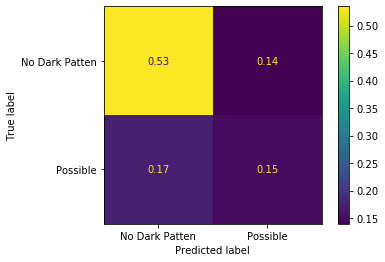}  
  \caption{Sneaking}
  \label{fig:confusion-sneaking}
\end{subfigure}
\begin{subfigure}{.45\textwidth}
  \centering
  \includegraphics[width=1\linewidth]{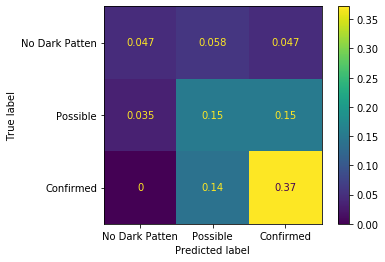}  
  \caption{Interface Interference}
  \label{fig:confusion-interface}
\end{subfigure}
\begin{subfigure}{.45\textwidth}
  \centering
  \includegraphics[width=1\linewidth]{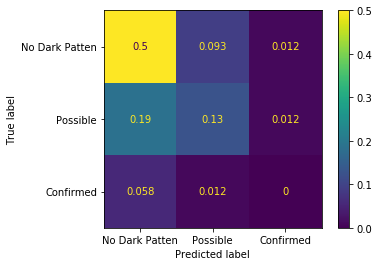} 
  \caption{Forced Action}
  \label{fig:confusion-forced}
\end{subfigure}

\caption{Confusion matrix of different dark patterns}
\label{fig:confusion}
\end{figure}

\clearpage

\section{Description of the data repository}\label{sec:repodescription}
The repository where the work is located contains the following resources
\begin{itemize}
    \item \emph{clusteringnotyesoption.ipynb} - The python notebook used for clustering of "Not Yes" options for the cookie banners. It is supposed to be uploaded and ran on Google Collab environment
    \item \emph{automatedanalysisofdarkpatternscookie.ipynb} - The python notebook used for performing parameter search and training a Gradient Boosted Classifier for the dataset. It is ran on local Jupyter Notebook environment.
    \item \emph{data->banner\_data\_clean.csv} - The cleaned dataset used in training of the classifier.
    \item \emph{data->banner\_data.csv} - The original data from the project it is not used for training.
    \item \emph{data->cookie\_consent.db} - The translated dataset for Privacy Policy and Cookie Policy documents.
    \item \emph{snippets} - The home for all the little snippets created to help in this project.
\end{itemize}

\end{document}